%% file: main.tex
\documentclass{article}

\usepackage[preprint]{icml2026}

\usepackage{lineno}
\usepackage[utf8]{inputenc} 
\usepackage[T1]{fontenc}    
\usepackage{url}            
\usepackage{booktabs}       
\usepackage{amsfonts}       
\usepackage{nicefrac}       
\usepackage{microtype}      
\usepackage{xcolor}         
\usepackage{lipsum}
\usepackage{wrapfig, booktabs}
\usepackage{booktabs}    
\usepackage{subcaption}  
\usepackage{tabularx}
\usepackage{booktabs} 
\usepackage{natbib}
\usepackage{pifont}
\usepackage{amsmath}
\usepackage{hyperref}
\usepackage{cleveref}

\input{math_commands.tex}
\input{preamble}

\icmlsetsymbol{equal}{*}
\icmlsetsymbol{corresponding}{†}

\begin{document}

\renewcommand{\icmlsetsymbol}[1]{}

\twocolumn[
  \icmltitle{Projected Representation Conditioning for High-fidelity Novel View Synthesis}

  \begin{icmlauthorlist}
    \icmlauthor{Minseop Kwak}{equal,yyy}
    \icmlauthor{Minkyung Kwon}{equal,yyy}
    \icmlauthor{Jinhyeok Choi}{equal,yyy}
    \icmlauthor{Jiho Park}{yyy}
    \icmlauthor{Seungryong Kim}{yyy}
  \end{icmlauthorlist}

  \icmlcorrespondingauthor{Seungryong Kim}{seungryong.kim@kaist.ac.kr}

  \icmlaffiliation{yyy}{KAIST AI}

  \begin{center}
    {\tt \href{https://cvlab-kaist.github.io/ReNoV}{https://cvlab-kaist.github.io/ReNoV}}
  \end{center}

  \vskip 0.3in
]

\printAffiliationsAndNotice{%
  \icmlEqualContribution
}

\icmlkeywords{Machine Learning, ICML}

\begin{abstract}
We propose a novel framework for diffusion-based novel view synthesis that leverages external representations as conditioning signals, exploiting their geometric correspondence properties and reconstruction capabilities to enhance geometric consistency and inpainting quality in generated novel viewpoints. Motivated by the observation that strong correspondence capabilities emerge within the spatial attention of external visual representations, we introduce a representation-guided novel view synthesis framework, dubbed \textbf{ReNoV} (\textbf{Re}presentation-guided \textbf{No}vel \textbf{V}iew synthesis), equipped with dedicated projection modules that inject external representations into the diffusion process. Extensive experiments demonstrate that our design significantly improves both reconstruction fidelity and inpainting quality, outperforming prior diffusion-based novel view synthesis methods on standard benchmarks and enabling robust synthesis from sparse, unposed image collections. 
\end{abstract}

\input{Writing/1_introduction}

\input{Writing/2_related_work}
\input{Writing/3_motivation}
\input{Writing/4_analysis}
\input{Writing/5_methodology}

\input{Writing/6_experiments}
\input{Writing/7_conclusion}

\input{appendix}
\bibliography{icml2026_conference}
\bibliographystyle{icml2026}

\end{document}

%% file: math_commands.tex
\usepackage{amsmath,amsfonts,bm}

\def\eqref#1{equation~\ref{#1}}

\def\1{\bm{1}}

\DeclareMathAlphabet{\mathsfit}{\encodingdefault}{\sfdefault}{m}{sl}
\SetMathAlphabet{\mathsfit}{bold}{\encodingdefault}{\sfdefault}{bx}{n}

%% file: preamble.tex
\usepackage{lipsum}

\usepackage{kotex}
\usepackage{xcolor,colortbl}
\usepackage{xspace}
\usepackage{adjustbox}
\usepackage{multirow}
\usepackage{graphicx}
\usepackage{amsmath}
\usepackage{subcaption}

\definecolor{tabfirst}{rgb}{1, 0.7, 0.7}
\definecolor{tabsecond}{rgb}{1, 0.85, 0.7}
\definecolor{tabthird}{rgb}{1, 1, 0.7}

%% file: Writing/1_introduction.tex
\section{Introduction}

Novel view synthesis—predicting scene appearance from target camera viewpoints—has long been a fundamental challenge in computer vision. Recent diffusion models enable novel view generation without explicit 3D representations such as Neural Radiance Fields~\citep{mildenhall2021nerf} or 3D Gaussian Splatting~\citep{kerbl20233d}. At the same time, diffusion-based novel view synthesis, including multi-view diffusion models~\cite{watson2022novel, liu2023zero, shi2023zero123++, gao2024cat3d, szymanowicz2025bolt3d}, leverage generative priors from large-scale 2D diffusion models~\cite{rombach2022high} to synthesize novel views. These approaches map reference and noisy target images into a shared feature space, enabling the model to generate target views consistent with the reference.


Maintaining consistency across reference views and between reference-target views is central to novel view synthesis. Recent visual foundation models~\cite{wang2025vggt, lin2025depth} trained for multi-view reasoning already possess such geometric and semantic correspondence abilities. We hypothesize that leveraging these powerful representations can serve as effective prompting signals for diffusion-based novel view synthesis. To ensure consistency between reference and target views, we draw inspiration from warping-and-inpainting novel view synthesis approaches and incorporate multi-view features into a network that projects reference features into 3D space and reprojects them onto the target viewpoint, explicitly bridging the coherence between reference and target views.

In our analysis, we observe that novel view synthesis requires multiple capabilities: faithful reconstruction of visible regions from reference viewpoint, plausible inpainting of regions occluded in the reference images, and appropriate locality for coherent object grouping. We conduct in-depth analysis of features~\citep{oquab2023dinov2, wang2025vggt, lin2025depth} regarding their semantic, geometric and locality awareness, as well as their novel view reconstruction capabilities from warped geometry. Our analysis reveals the geometry-enhancing capabilities of external representations, especially VGGT~\citep{wang2025vggt} and DA3~\cite{lin2025depth}, whose rich, geometrically multi-view consistent features make them suitable for novel view conditioning and generation from multiple reference images.

In this light, we introduce a novel framework, named \textbf{Re}presentation-guided \textbf{No}vel \textbf{V}iew synthesis (shortened \textbf{ReNoV}) that leverages powerful features for novel view image prediction. We design a multi-view synthesis architecture where a reference network extracts features from multiple source views, which are then aggregated with the target-view generation features via attention in a generation network. To enhance reconstruction and inpainting performance at target viewpoint generation, we introduce projected representation conditioning, a generalizable method that geometrically warps reference view external representations to the novel viewpoint, providing condition to improve the diffusion model's synthesis quality. This approach enables our model to generate high-fidelity novel views while maintaining 3D consistency across diverse scenes and camera transformations.

Extensive experiments on RealEstate10K benchmark~\cite{zhou2018stereo} and zero-shot evaluation on DTU benchmark~\cite{jensen2014large} demonstrate that our method shows competitive results to state-of-the-art feedforward novel view synthesis approaches across both interpolation and extrapolation settings, with ablation studies confirming the effectiveness of our integrated semantic and geometric conditioning approach.

%% file: Writing/2_related_work.tex
\section{Related work}
\label{Sec: related_work}

\paragraph{Diffusion-based 3D generation models.}
\label{Sec: nvs-not-generative}
Prior efforts in generative 3D and multi‐view synthesis have largely focused on leveraging diffusion models to bridge the gap between 2D image priors and 3D scene representations. DreamFusion~\citep{poole2022dreamfusion} first demonstrated text‐to‐3D generation by optimizing a Neural Radiance Field with a pretrained 2D diffusion prior, while ProlificDreamer~\citep{wang2023prolificdreamer} extended this paradigm by distilling multi‐view diffusion signals into a feed‐forward geometry network for faster inference. In the multi‐view setting, MVDream~\citep{shi2023mvdream} proposes a view‐consistent denoising pipeline that jointly refines color and depth across posed images, and Zero123~\citep{liu2023zero} tackles single‐image to novel‐view synthesis via a conditioned diffusion model that hallucinates plausible viewpoints. ZeroNVS~\citep{zeronvs} extends upon this method for single-view novel view synthesis, while CAT3D~\citep{gao2024cat3d} employs spatial cross-attention between generating viewpoints to achieve consistent novel view synthesis at target viewpoints. ViewCrafter~\citep{yu2024viewcrafter} delivers high-fidelity performance by finetuning a video diffusion model that conditions on point cloud representations reconstructed from the input images, enabling precise camera pose control through explicit 3D geometric priors. To handle large viewpoint changes, ViewCrafter employs an iterative view synthesis strategy with camera trajectory planning to progressively expand the point cloud coverage and synthesize novel views in previously occluded regions. While these methods have achieved impressive visual quality, they either require costly per‐scene optimization or video multi-frame generation, or rely on known camera poses, struggling with large pose extrapolation.

\paragraph{Feedforward 3D regression models.}
Feed‐forward approaches to novel‐view synthesis and 3D reconstruction bypass costly per‐scene optimization by learning rich geometric priors from large‐scale training. PixelNeRF~\citep{yu2021pixelnerf} first demonstrated how to condition a NeRF on input views via local CNN features, and IBRNet~\citep{wang2021ibrnet} built on this by fusing multi‐view depth and appearance cues in a self‐supervised stereo framework. MVSplat~\citep{chen2024mvsplat} further refines this paradigm by estimating 3D Gaussians through cost‐volume–based depth prediction, achieving high‐fidelity volumetric representations from sparse inputs. Concurrently, single‐image methods like ShapeFormer~\citep{yan2022shapeformer} exploit transformer architectures to generate novel views and coarse geometry from a one-shot image. More recently, transformer‐based systems such as DUSt3R~\citep{dust3r_cvpr24} and MASt3R~\citep{leroy2024grounding} have learned to predict point‐maps and camera poses directly from unposed images, while Noposplat~\citep{ye2024no} unifies pose estimation with 3D Gaussian fitting in a single feed‐forward pass. Likewise, FLARE~\citep{zhang2025flare} proposes a cascaded feed-forward approach that uses camera pose estimation as a bridge to guide subsequent geometry reconstruction and appearance learning for sparse-view novel view synthesis, and AnySplat~\citep{jiang2025anysplat} predicts novel view images as well as Gaussian primitives from uncalibrated image collections. However, despite their efficiency, these feed‐forward models remain fundamentally limited by reference‐view visibility, often failing to extrapolate to unseen angles or complete occluded structures without explicit inpainting.

\paragraph{Geometry prediction models.}  
Recent advances in geometry prediction models have enabled powerful geometric reasoning from sparse image inputs. VGGT~\citep{wang2025vggt} and DepthAnythingV3~\citep{lin2025depth} are state-of-the-art approaches capable of predicting camera parameters, depth maps, and point maps from a set of unposed images. Notably, both models build upon DINOv2~\citep{oquab2023dinov2}, inheriting its rich prior, capturing both semantic and geometric structure through self-supervised learning, yet they adopt different architectural strategies. VGGT extracts DINOv2 features as input and processes them through a separate transformer network with alternating frame attention and global attention. In contrast, Depth Anything V3 directly fine-tunes the DINOv2 model, selectively applying either frame attention or global attention at each layer. Both models then pass selected intermediate features through DPT networks~\cite{ranftl2021vision}, decoding the features into depth, point map, and camera parameters. VGGT additionally estimates point tracks for the input images, enabling higher-order geometric reasoning. While two models differ in architecture and supervision, both descend from DINOv2 and learn to reason across multiple views for geometry prediction.
In this work, we analyze how these differences affect their representations and leverage them as signals for diffusion-based novel view synthesis.

%% file: Writing/3_motivation.tex



%% file: Writing/4_analysis.tex
\section{Motivation and Analysis}
\label{Sec :Analysis}
\begin{figure}[t!]
  \centering
  \includegraphics[width=\linewidth]{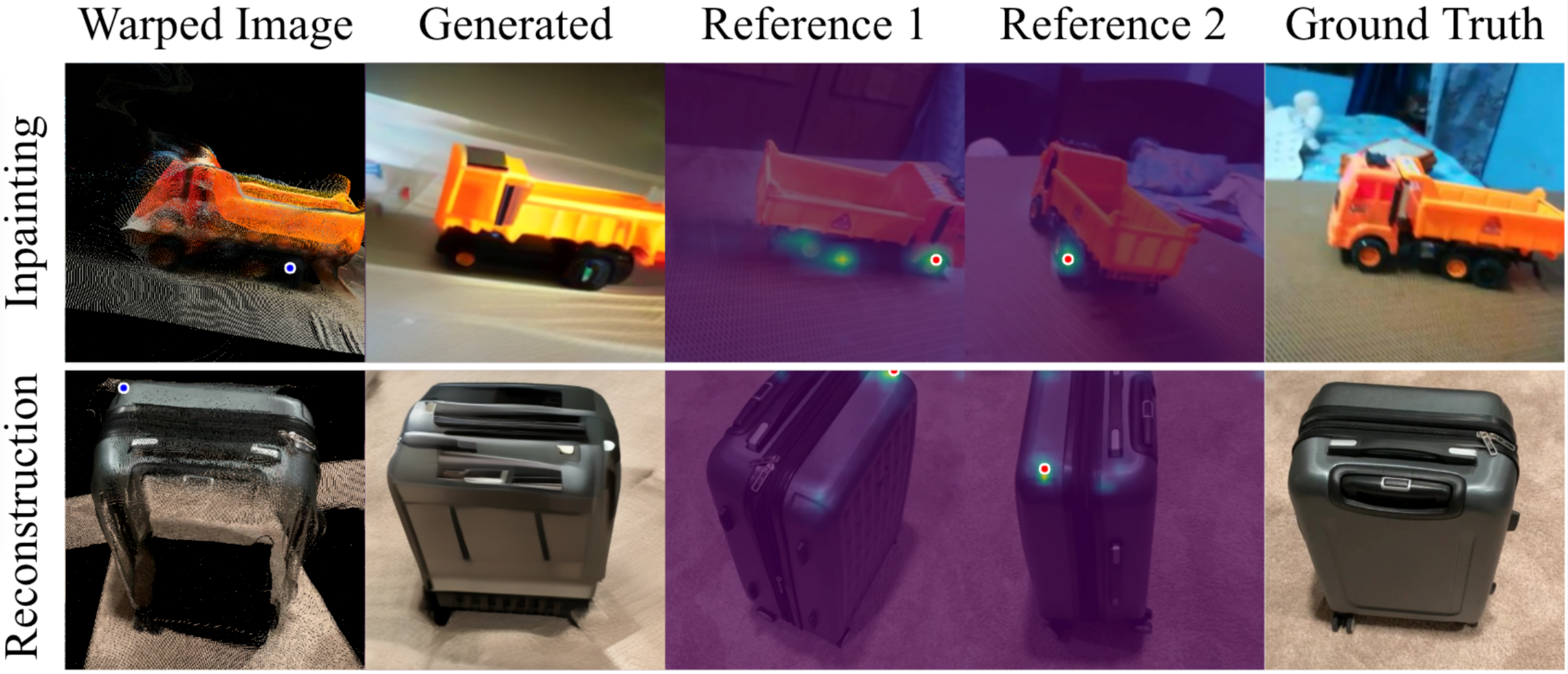}
  \caption{
    \textbf{Cross-view attention maps of the denoising network~\citep{seo2024genwarp, kwak2025aligned}.} 
    A query pixel (blue dot) is chosen in the warped target view, and the resulting 
    cross-attention weights on two reference images are visualized. \textbf{Inpainting}: 
    the wheel is absent in the warped view, so attention shifts to the corresponding 
    wheels in the references. \textbf{Reconstruction}: the suitcase edge is visible, so attention 
    concentrates on the geometrically aligned edges to refine the reconstruction.
  }
  \label{fig:attention_vis}
  \vspace{-10pt}
\end{figure}

\begin{figure*}[t!]
  \centering
  \begin{subfigure}[b]{0.33\linewidth}
    \includegraphics[width=\linewidth]{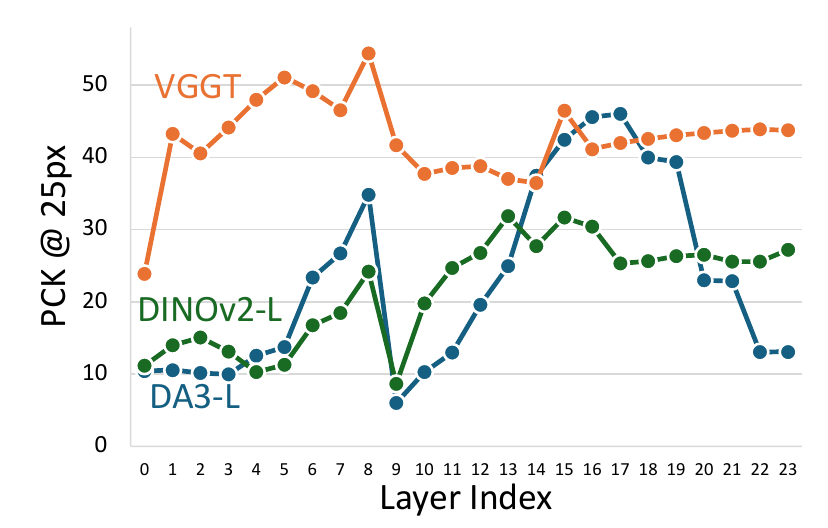}
    \caption{Geometric correspondence across layers}
    \label{fig:main_analysis_a}
  \end{subfigure}\hfill
  \hspace{-5px}
    \begin{subfigure}[b]{0.33\linewidth}
    \includegraphics[width=\linewidth]{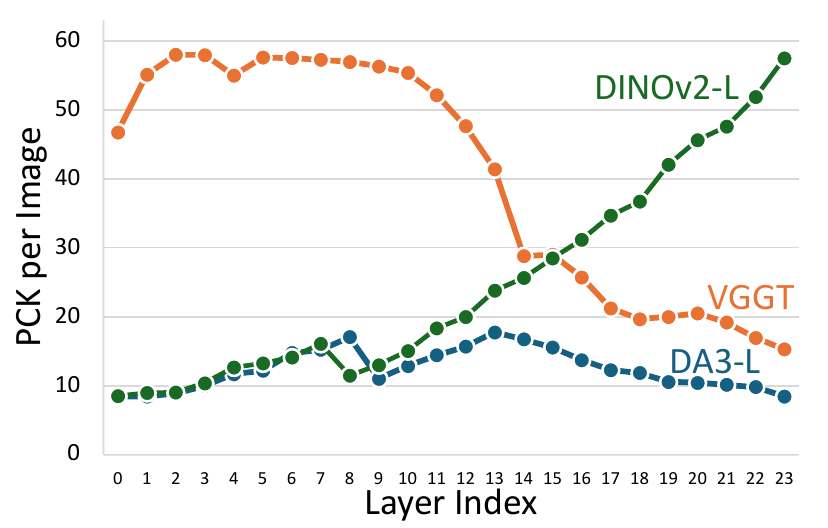}
    \caption{Semantic correspondence across layers} 
    \label{fig:main_analysis_b}
  \end{subfigure}
  \hspace{-5px}
  \begin{subfigure}[b]{0.33\linewidth}
    \includegraphics[width=\linewidth]{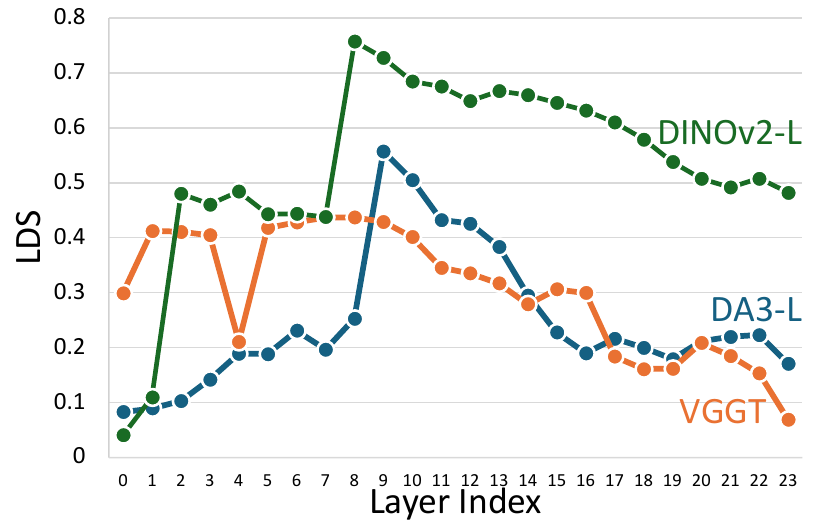}
    \caption{Local vs. Distant Similarity} 
    \label{fig:main_analysis_c}
  \end{subfigure}
  \vspace{-5pt}
  \caption{\textbf{Analysis of visual foundation models.} (a) Geometric correspondence, (b) Semantic corrspondence \& (c) Local vs. Distant Similarity across feature layers in VGGT~\citep{wang2025vggt}, DA3-Large~\citep{lin2025depth} \& DINOv2-Large~\citep{oquab2023dinov2}. }
  \label{fig:main_analysis}
\end{figure*}

\begin{figure*}[t]
  \centering
  \vspace{-10pt}
  \includegraphics[width=\textwidth,]{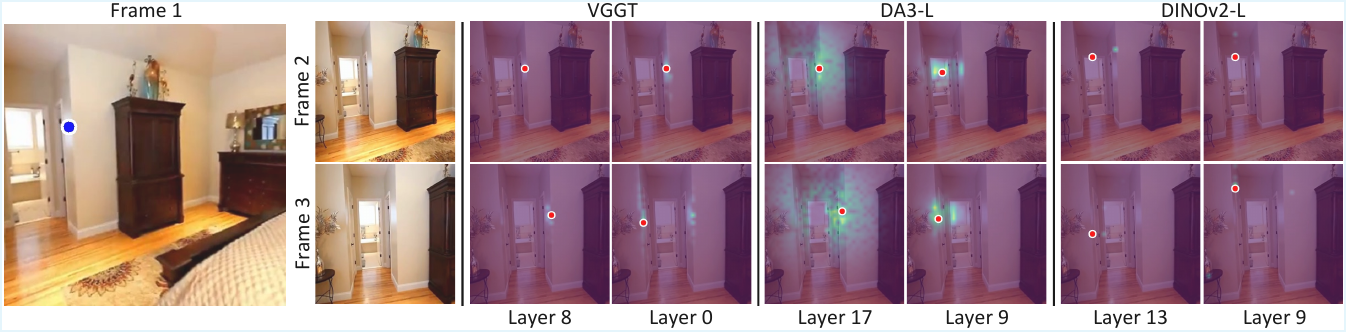}
  \vspace{-18pt}
  \caption{
      \textbf{Geometric correspondence.} A query point (blue dot) is selected in Frame 1, and cosine similarity maps are computed in Frame 2 and Frame 3. The scene contains featureless walls, allowing assessment of whether the model can localize the geometrically corresponding instance. Deeper layers of VGGT and DA3-L accurately identify the correct location in the corner wall aligned with the query point, while early layer 0 of VGGT and the feature of DINOv2 attend to incorrect but semantically similar locations in the wall. This illustrates that deeper layers of VGGT and DA3-L capture geometric structure more reliably than others.
  }
  \label{fig:geo_corr}
\end{figure*}

\input{Tables/recon_quan}

As discussed in Sec.~\ref{Sec: related_work}, novel view synthesis approaches fall into several categories. Non-generative approaches—e.g., MVSplat\citep{chen2024mvsplat} and NopoSplat~\citep{ye2024no}—do not exploit generative models and therefore cannot infer geometry or appearance in regions unseen or occluded in the reference images. In contrast, diffusion-based generative methods can extrapolate to viewpoints distant from the inputs; however, as these methods condition the diffusion models on target camera pose as a feature embedding, they remain confined to the pose distribution encountered during training, precluding truly arbitrary novel‐pose synthesis.

We interpret novel view synthesis as a warping-and-inpainting problem, akin to GenWarp~\citep{seo2024genwarp} and MoAI~\citep{kwak2025aligned}, requiring models to excel at two tasks: accurate \textit{reconstruction} of visible regions and consistent \textit{inpainting} of occluded regions. Within diffusion-based frameworks, both reconstruction and inpainting are achieved by implicitly aggregating features from reference viewpoints through the U-Net's spatial attention modules, driven by conditioning features that establish cross-view correspondences. This naturally leads to the question: what properties should an ideal conditioning feature possess for effective novel view generation?

To this end, we examine diffusion‐model attention during novel‐view synthesis and uncover a consistent pattern (Fig.~\ref{fig:attention_vis}): regions visible in the reference views—those requiring reconstruction—attend sharply to their geometric correspondences, whereas regions needing inpainting attend broadly to semantically similar locations in the references. This can be intuitively understood, as reconstruction performance hinges on pinpointing exact correspondences, while inpainting relies on semantically related context to synthesize unseen areas coherently. This motivates the search for a conditioning representation that simultaneously encodes semantic awareness and geometric correspondence. In the next section, we evaluate several representations~\citep{oquab2023dinov2, he2022masked, wang2025vggt, lin2025depth} to identify the representation that best balances semantic awareness with geometric correspondence, and offer a comprehensive analysis. To identify the optimal conditioning feature for our warping-and-inpainting diffusion framework, we compare several widely-used representations—DINOv2-L~\citep{oquab2023dinov2}, VGGT~\citep{wang2025vggt}, and DepthAnything V3-L~\citep{lin2025depth}.

\paragraph{Correspondence capabilities.}
To assess the geometric correspondence capabilities of various representations, we qualitatively and quantitatively evaluate cross-view similarity for intermediate features of each model, as shown in Fig.~\ref{fig:main_analysis} and Fig.~\ref{fig:geo_corr}. In Fig.~\ref{fig:main_analysis}, we provide per-layer quantitative results of geometric correspondence between features from multi-view images, measuring three different metrics: in Fig.~\ref{fig:main_analysis_a}, we provide geometric correspondence values, in Fig.~\ref{fig:main_analysis_b}, semantic correspondence values, and lastly, in Fig.~\ref{fig:main_analysis_c} the local vs. distant similarity (LDS) metric propsed in iREPA~\cite{singh2025matters} for measuring spatial self-similarity. In Fig.~\ref{fig:geo_corr}, we present a qualitative visualization of similarity maps for a triplet of multi-view images from a single scene, where a query point is selected in the first frame, and similarity maps are computed by comparing the first frame's features with those of the second and third frames. 

The qualitative values reveal that DINOv2~\citep{oquab2023dinov2} frequently fails to disambiguate repeated structures, revealing a lack of geometric awareness. For the VGGT~\citep{wang2025vggt} representation, we find that deeper layers (8 and on onwards) effectively capture geometric structure, attending to the correct location in the corner of the wall that is spatially aligned with the query point in subsequent frames - the quantitative results show similar results in the latter layers, showing stable and consistent geometric correspondence performance. The intermediate features of DepthAnythingV3-Large (DA3-L)~\citep{lin2025depth} exhibit progressively stronger geometric correspondence in deeper layers, reaching peak performance at layer 17. We observe that VGGT and DA3-L process multiple frames jointly and leverage their global attention mechanisms to capture geometric structure consistently across views, enabling precise localization of the corresponding object instance even in the presence of repeated or ambiguous patterns.


\paragraph{Representation reconstruction capabilities.}
Building on our geometric correspondence analysis, we evaluate the intermediate features' reconstruction and inpainting capabilities for novel viewpoints through novel-view projection, examining how these correlate with their geometric correspondence metrics. To this end, we train a shallow MAE~\citep{he2022masked} decoder to predict a target view image from the warped projection of the reference view image features. The optimal feature representation should encapsulate multi-view semantic and geometric information, enabling the model to accurately reconstruct visible regions while effectively inpainting occluded areas.

\begin{figure}[t]
    \begin{center}
        \includegraphics[width=\linewidth]{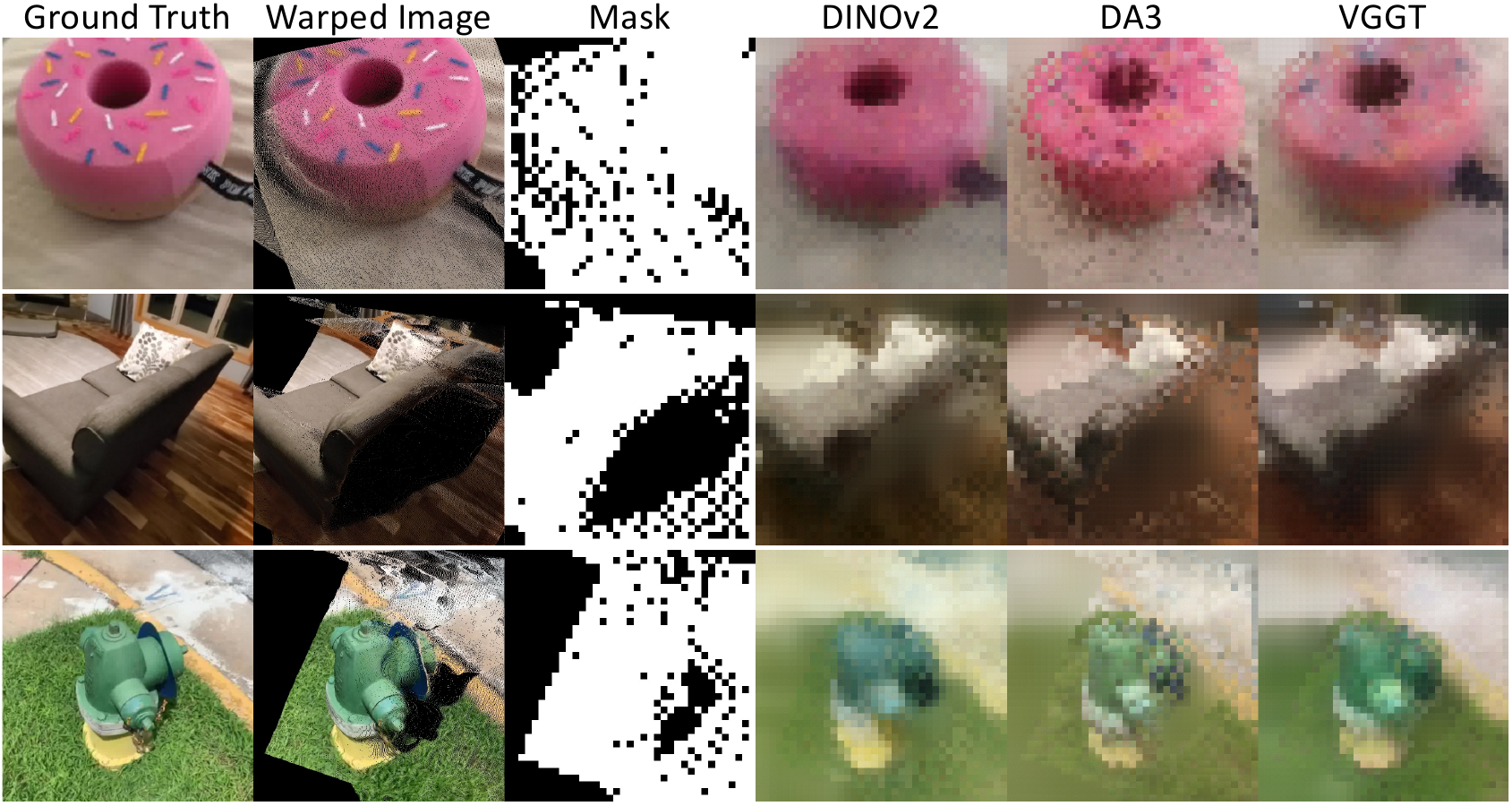}
    \end{center}
    \vspace{-10pt}
    \caption{
        \textbf{Qualitative results for feature reconstruction analysis.}
        We warp the extracted features using point clouds,
        resulting in feature-level holes that require inpainting.
    }
    \vspace{-15pt}
    \label{fig:recon_qual}
\end{figure}
For DINOv2, we directly probe the encoder output, whereas for VGGT and DA3-L, we extract intermediate features~\footnote{layer 4, 11, 17, 23 for VGGT.} \footnote{layer 11, 15, 19, 23 for DA3-L.} and use all of them in our analysis. To facilitate feature warping, we employ an off-the-shelf geometry prediction model~\citep{wang2025vggt} to obtain the pointmaps and camera poses. The token-level features are re-projected into the target view; patches without valid projections are replaced by learnable mask tokens, and training is supervised with a mean-squared-error objective. For quantitative results, we evaluate each model for different numbers of reference views using PSNR, SSIM, and LPIPS metrics. Table~\ref{tab:recon} shows that VGGT features consistently achieve the highest results across all metrics and inference settings. In the qualitative results, Fig.~\ref{fig:recon_qual} also demonstrates that the generated images using VGGT features are most visually accurate compared to the target view images. 

The quantitative and qualitative results reveal a strong correlation between the geometric correspondence capability and reconstruction performance. As shown in Table~\ref{tab:recon}, VGGT features generally achieve the highest scores across all metrics and reference view configurations, aligning with their superior geometric correspondence in deeper layers (Fig.\ref{fig:main_analysis_a}). In contrast, DINOv2's limited geometric correspondence capability, as shown in per-layer analysis, translates to inferior reconstruction quality in both quantitative metrics and qualitative results, as visualized in Fig.~\ref{fig:recon_qual}. These findings imply that geometric correspondence is an important correlating factor regarding the reconstructive capability of a visual representation, as precise spatial alignment is essential for both accurately reconstructing visible regions and coherently inpainting occluded areas from a geometrically consistent reference context.


%% file: Tables/recon_quan.tex
\begin{table*}[t]
  \centering
  \small
  \resizebox{0.95\linewidth}{!}{%
  \begin{tabular}{l|ccc|ccc|ccc}
      \toprule
      \multirow{2}{*}{\textbf{Model}} & \multicolumn{3}{c}{PSNR $\uparrow$} & \multicolumn{3}{|c}{SSIM $\uparrow$} & \multicolumn{3}{|c}{LPIPS $\downarrow$} \\
      \cmidrule(lr){2-4} \cmidrule(lr){5-7} \cmidrule(lr){8-10}
      & 1 view & 2 view & 3 view & 1 view & 2 view & 3 view & 1 view & 2 view & 3 view \\
      \midrule
      Warped image & 8.30 & 11.33 & 11.77 & 0.086 & 0.206 & 0.228 & 0.770 & 0.713 & 0.705 \\
      DA3~\cite{lin2025depth} & 15.43 & 17.20 & 17.13 & 0.515 & 0.517 & 0.511 & 0.751 & 0.747 & 0.749 \\
      DINOv2-L~\cite{oquab2023dinov2} & 15.56 & 16.76 & 16.63 & \textbf{0.548} & \textbf{0.554} & \textbf{0.553} & 0.763 & 0.757 & 0.759 \\
      VGGT~\cite{wang2025vggt} & \textbf{15.59} & \textbf{17.30} & \textbf{17.22} & 0.539 & 0.549 & 0.547 & \textbf{0.733} & \textbf{0.718} & \textbf{0.722} \\
      \bottomrule
  \end{tabular}}
  \vspace{5pt}
  \caption{
      \textbf{Quantitative evaluation of feature analysis.} We evaluate reconstruction capability of each feature across reference view counts.
  }
  \vspace{-10pt}
  \label{tab:recon}
\end{table*}


%% file: Writing/5_methodology.tex
\section{Method}
\subsection{Overview}

Our objective is to predict a novel view image $I_\text{tgt}$ for target viewpoint $\pi_\text{tgt}$ by leveraging both the generative capabilities of diffusion models and the semantic-geometric correspondence of external feature representations validated in our analysis. Given $N$ unposed and sparse reference images $\mathcal{I_\text{ref}} = \{ I_n \in \mathbb{R}^{H \times W \times 3} \}_{n=1}^N$, we adopt a dual U-Net architecture following \citep{seo2024genwarp}, reminiscent of ControlNet~\citep{zhang2023adding}. The reference U-Net extracts multi-view features by processing input images alongside their conditioning geometric information and representations from external models~\citep{oquab2023dinov2, wang2025vggt, lin2025depth}, while the denoising U-Net synthesizes the target view through iterative refinement of a noisy latent, conditioned features from the reference network as well as geometrically warped external features of the reference images.

\subsection{Reference conditioning}
\paragraph{Geometry conditioning. }
We begin by leveraging an off-the-shelf geometry prediction model~\citep{wang2025vggt} to estimate a set of camera poses \(\{ \pi_n \in \mathbb{R}^{4 \times 4} \}_{n=1}^{N}\) and corresponding pointmaps \(\{ P_n \in \mathbb{R}^{H \times W \times 3} \}_{n=1}^{N}\), where each \(P_n\) is a 2D grid of 3D points representing the predicted world coordinates for the pixels of the reference image \(I_n\). To incorporate geometric priors into our model, we apply a positional embedding function \(\gamma(\cdot)\) to each pointmap, resulting in Fourier-encoded features \(\gamma(P_n)\), which is passed through a small pose guider network~\cite{hu2024animate} to be used as a condition for the reference and geometry prediction network.

\begin{figure*}[t]
\centering
    \includegraphics[width=\textwidth]{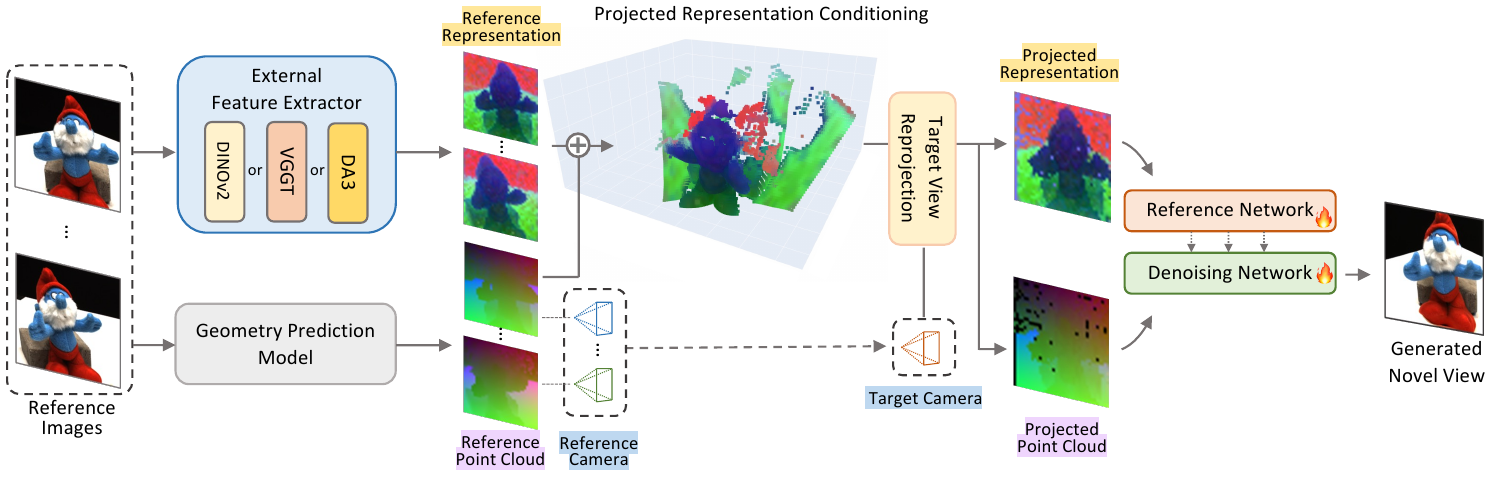}
    \vspace{-20pt}
    \caption{\textbf{Model architecture.} Given $N$ reference images, we extract visual features, dense point clouds, and camera poses using an external representation model (e.g., VGGT~\cite{wang2025vggt}, DA3~\cite{lin2025depth}, or DINOv2~\cite{oquab2023dinov2}). These components undergo projected representation conditioning, where reference features and point clouds are projected into the target camera frustum to form warped representation and point-map planes. The reference network aggregates these multi-view inputs by passing them as keys and values to denoising network. Simultaneously, the denoising network receives the projected feature and point cloud planes as direct conditioning, aggregating reference cues to synthesize the novel view image.}
    \label{fig:architecture}
\vspace{-5pt}
\end{figure*}

\paragraph{Representation conditioning.}
We begin by jointly processing the $N$ reference images through an external model to extract their feature representations, which are then warped to the target viewpoint to condition the denoising network—a process we term \textit{projected representation conditioning}. For DINOv2, we use the final layer features for representation conditioning. For VGGT and DepthAnythingV3, we extract intermediate features from transformer layers and concatenate them, following their original architectural designs. For each reference image \(I_n\), we obtain both local and global features at each selected layer, denoted as \(t_{l,n}\) and \(t_{g,n} \in \mathbb{R}^{H/P \times W/P \times 1024}\), respectively. These features are concatenated along the channel dimension to form a unified representation \(T_n = [t_{g,n}; t_{l,n}] \in \mathbb{R}^{H/P \times W/P \times 2048}\). This obtained feature is high-dimensional, exceeding what the reference U\textendash Net can efficiently process. To address this, we reduce the channel dimensionality of \(T_n\) through a convolutional network, and concatenate it with the geometric conditioning before incorporation in the initial features of reference U\textendash Nets.


Analogous to the geometric conditioning, we apply a positional embedding function \(\gamma(\cdot)\) to the extracted features \(T_n\), resulting in the Fourier-encoded representation \(\gamma(T_n)\). The final reference condition \(c_n\) is obtained by concatenating the encoded image features and pointmaps:
\[
c_n = [\gamma(P_n) ; \gamma(T_n)].
\]
Following the approach of Hu et al.~\citep{hu2024animate}, each condition vector \(c_n\) is passed through a shallow convolutional network and then added to the image latents prior to input to the reference U\textendash Net.

\subsection{Projected representation conditioning}
To enhance the fidelity of reconstruction and inpainting in novel view synthesis, we incorporate a geometry-driven conditioning mechanism based on warping. Specifically, we project the reference pointmaps \(\{P_1, \dots, P_N\}\) and the corresponding external representation features \(\{T_1, \dots, T_N\}\) into the target viewpoint \(\pi_{\text{tgt}}\). These projected signals provide spatial priors that guide the diffusion model toward higher-quality generation results. First, the set of reference pointmaps \(\{P_1, \dots, P_N\}\), expressed in a global coordinate frame, can be directly aggregated to form a unified point cloud:$\mathcal{P}_{\text{ref}}$. This point cloud \(\mathcal{P}_{\text{ref}} \in \mathbb{R}^{(N \times H \times W) \times 3}\) is then projected onto the target viewpoint \(\pi_{\text{tgt}}\):
\begin{equation}
  \mathcal{P}^{\Pi}_{\text{tgt}} = \Pi(\mathcal{P}_{\text{ref}}, \pi_{\text{tgt}}).
 \label{equation:projection}
\end{equation}
When multiple points are projected to the same pixel, only the one closest to the target image plane is retained, following the standard point cloud rasterization procedure~\citep{seo2024genwarp}. The resulting projected pointmap \(\mathcal{P}^{\Pi}_{\text{tgt}}\) serves as a sparse geometric condition that guides the generation of \(I_{\text{tgt}}\) from the reference views.

Given the observed multiview-consistent nature of geometric~\citep{wang2025vggt, lin2025depth} external representations, we unproject them into 3D space by anchoring each pixel-level feature to its corresponding 3D coordinate from the predicted pointmap \(P_n \in \mathbb{R}^{H \times W \times 3}\), forming a 3D feature point cloud. This pointcloud is then projected into the target view, yielding a spatially aligned warped feature map. The projected features \(T_{\text{tgt}}^{\Pi}\) and projected pointmap \(X_{\text{tgt}}^{\Pi}\) are provided as input conditions to the denoising network.

Following the same design as in the reference network, we encode \(X_{\text{tgt}}^{\Pi}\) and \(T_{\text{tgt}}^{\Pi}\) using a positional embedding function \(\mathcal{\gamma}(\cdot)\), and concatenate their Fourier embeddings with a binary visibility mask \(M_{\text{tgt}}\), which indicates grid pixels where no 3D point was projected. This forms the target correspondence condition \(c_{\text{tgt}}^d\):
\begin{equation}
c_{\text{tgt}}^d = [\mathcal{\gamma}(X_{\text{tgt}}^{\Pi}), \mathcal{\gamma}(T_{\text{tgt}}^{\Pi}), M_{\text{tgt}}].
\end{equation}
The condition \(c_{\text{tgt}}^d\) is then processed by a shallow convolutional network and added to the noise latent before being passed into the denoising U\textendash Net. As discussed in Sec.~\ref{Sec :Analysis}, providing the warped feature \(T_{\text{tgt}}^{\Pi}\) to the denoising U\textendash Net serves two key purposes: it supplies semantic priors for unseen or occluded regions by leveraging multiview-consistent features, and it delivers accurate geometric information for regions visible in the reference views. This conditioning enables the model to generate more structurally faithful outputs at the target view \(\pi_{\text{tgt}}\).

\subsection{Novel-view image generation}
Following this, we conduct integrated self-and-cross attention between reference and target features, allowing the model to leverage other viewpoints, similar to \citep{seo2024genwarp}. Specifically, from the denoising U-Net, we extract key and value features of the target view, \(F_{\text{tgt}}^k, F_{\text{tgt}}^v \in \mathbb{R}^{1 \times C \times (W \times H)}\), obtained from spatial self-attention layers.  from spatial self-attention layers. These are concatenated along the viewpoint dimension with key and value features from $N$ reference views, so that the query feature $\mathbf{q} = F_{\text{tgt}}^q,$ is aggregated over attention map acquired with expanded key feature $\mathbf{k} = [F_{\text{tgt}}^k,\ F_1^k,\ \ldots,\ F_N^k]$ and value feature  $\mathbf{v} = [F_{\text{tgt}}^k,\ F_1^v,\ \ldots,\ F_N^k]$. where \(\mathbf{k}, \mathbf{v} \in \mathbb{R}^{(N+1) \times C \times (W \times H)}\). The aggregated attention is then computed as:
\begin{equation}
\text{Attention}(\mathbf{q}, \mathbf{k}, \mathbf{v}) = \text{softmax} \left( \frac{\mathbf{q}\mathbf{k}^T}{\sqrt{d_k}} \right) \mathbf{v}, \tag{6}
\end{equation}
where \(d_k\) denotes the dimensionality of the key features. Through this architecture, the generating U-Net can leverage features extracted from reference networks via attention aggregation, enabling NVS from multiple viewpoints. 




%% file: Writing/6_experiments.tex
\section{Experiments}

\subsection{Implementation details}


\newcommand{\cmark}{\ding{51}} 
\newcommand{\xmark}{\ding{55}} 

\begin{table*}[t]
  \centering
  \small
  \setlength{\tabcolsep}{6pt}
  \resizebox{0.9\linewidth}{!}{%
  \begin{tabular}{c|l|c|ccc|ccc}
    \toprule
    \multirow{2}{*}{\centering\textbf{Views}} & \multirow{2}{*}{\centering\textbf{Method}} & \multirow{2}{*}{\centering{\textbf{Pose-free}}}
      & \multicolumn{3}{c}{\textbf{Far-view Setting}}
      & \multicolumn{3}{c}{\textbf{Near-view Setting}} \\
    \cmidrule(lr){4-6}\cmidrule(lr){7-9}
     &  &  & PSNR$\uparrow$ & SSIM$\uparrow$ & LPIPS$\downarrow$
               & PSNR$\uparrow$ & SSIM$\uparrow$ & LPIPS$\downarrow$ \\
    \midrule
    \multirow{9}{*}{\centering\textbf{2-view}}
      & PixelSplat~\citep{charatan2024pixelsplat} & \xmark
      & 13.03 & 0.486 & 0.414
      & 11.57 & 0.330 & 0.634 \\
      & MVSplat~\citep{chen2024mvsplat} & \xmark
      & 12.22 & 0.416 & 0.423
      & 13.94 & 0.473 & 0.385 \\
      & NoPoSplat~\citep{ye2024no} & \cmark
      & 11.43	& 0.335	& 0.599
      & 11.44 & 0.357 & 0.576 \\
      & FLARE~\citep{zhang2025flare}& \xmark
      & 13.52 & 0.407 & 0.525 
      & 13.25 & 0.381 & 0.502 \\
      & LVSM~\citep{jin2024lvsm} & \xmark
      & \underline{15.23} & 0.499 & 0.415
      & \textbf{15.82} & 0.528 & 0.346 \\
      \cmidrule(lr){2-9}
      & \textbf{ReNoV} w/ DINOv2~\citep{oquab2023dinov2} & \cmark
      & 15.13 & \textbf{0.599} & 0.304
      & 14.70 & \textbf{0.602} & 0.277 \\
      & \textbf{ReNoV} w/ VGGT~\citep{wang2025vggt} & \cmark
      & \textbf{15.45} & \underline{0.584} & \textbf{0.297}
      & \underline{15.38} & \underline{0.599} & \underline{0.274} \\
      & \textbf{ReNoV} w/ DA3~\citep{lin2025depth} & \cmark
      & 14.38 & 0.550 & \underline{0.303}
      & 14.91 & 0.593 & \textbf{0.259} \\
    \midrule
    \midrule
    \multirow{6}{*}{\centering\textbf{1-view}}
      & LucidDreamer~\citep{chung2023luciddreamer} & \cmark
      & 12.96 & 0.248 & 0.385
      & 12.09 & 0.481 & 0.419 \\
      & GenWarp~\citep{seo2024genwarp} & \cmark
      & 8.69 & 0.253 & 0.597
      & 9.54 & 0.298 & 0.538 \\
      & ViewCrafter~\citep{ma2025efficient} & \cmark & 14.04 & 0.390 & 0.332 & 13.59 & 0.382  & 0.486 \\
      \cmidrule(lr){2-9}
      & \textbf{ReNoV} w/ DINOv2~\citep{oquab2023dinov2} & \cmark
      & \textbf{15.02} & \textbf{0.579} & \underline{0.328}
      & \textbf{14.13} & \textbf{0.574} & \underline{0.304} \\
      & \textbf{ReNoV} w/ VGGT~\citep{wang2025vggt} & \cmark
      & 14.08 & \underline{0.536} & 0.355
      & 13.91 & 0.542 & 0.333 \\
      & \textbf{ReNoV} w/ DA3~\citep{lin2025depth} & \cmark
      & \underline{14.35} & 0.534 & \textbf{0.325}
      & \underline{14.12} & \underline{0.550} & \textbf{0.292} \\
    \bottomrule
  \end{tabular}}
    \caption{\textbf{Zero-shot evaluation on the DTU~\citep{jensen2014large} dataset.} NoPoSplat and LVSM apply camera pose optimization in their test time. For a fair comparison, all models are evaluated in feed-forward manners, without test-time optimization. For further details, see ~\ref{supp:eval_detail}. \textbf{Bold} indicates the best performance, and \underline{underline} indicates the second best.}
    \label{table:dtu_zeroshot}
  \vspace{-10pt}
\end{table*}

For the image synthesis pipeline, we initialize from the pre-trained Stable Diffusion 2.1~\citep{rombach2022high}. The reference feature extraction networks share an identical architecture with the denoising U-Net but exclude timestep embeddings, as they are designed solely for semantic feature extraction rather than denoising operations.
Training is conducted on three multi-view datasets: RealEstate10K~\citep{zhou2018stereo} for diverse indoor/outdoor scenes, Co3D~\citep{reizenstein2021common} for object-centric captures, and MVImgNet~\citep{yu2023mvimgnet} for extensive multi-view imagery. We generate pseudo ground-truth geometry with an external geometry predictor~\citep{wang2025vggt, lin2025depth}, which provides both depth maps and normal predictions to establish reliable geometric supervision. During training, reference pointmaps is leveraged for explicit geometric warping of external representation between viewpoints and establishment of geometric conditioning signals that guide generation. The external representations undergo geometry-aware warping, ensuring proper transfer of spatial and semantic information across viewpoints while maintaining geometric consistency throughout synthesis.

\begin{figure*}[ht!]
    \centering
    \includegraphics[width=1.0\textwidth]{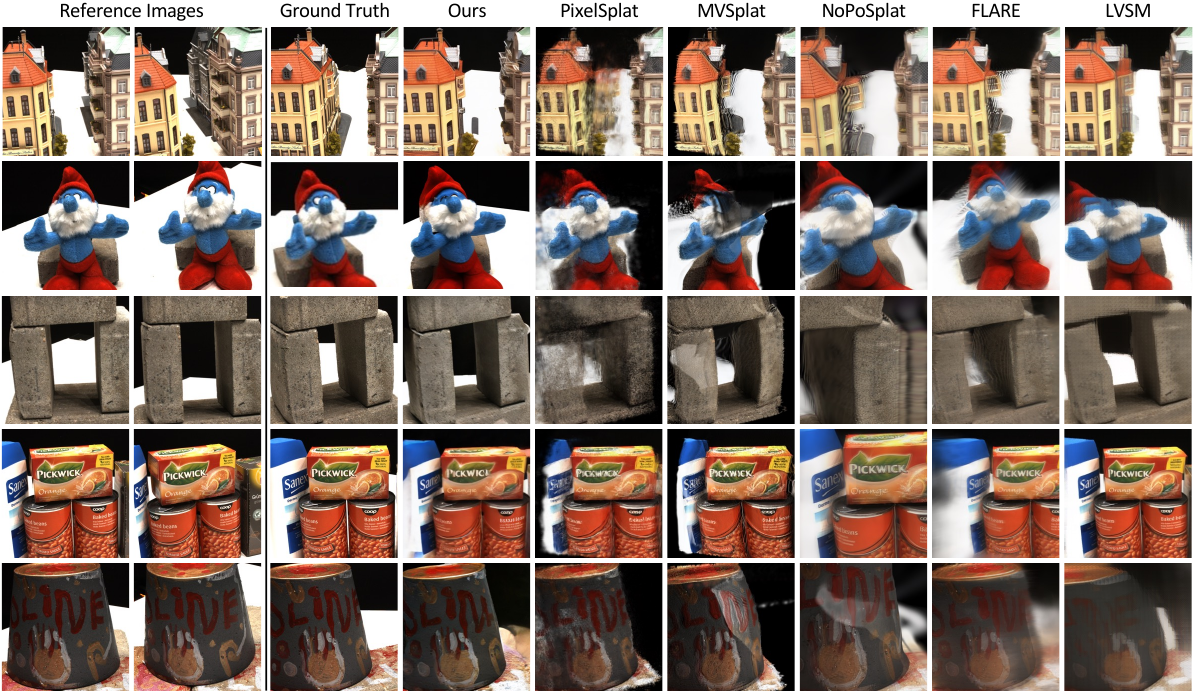}

    \caption{\textbf{Qualitative comparison on far-view setting at DTU dataset.} 
    Qualitative results of our model using far-view camera setting demonstrate our model (ReNoV w/ VGGT)'s extrapolative capabilities to plausibly generate locations not seen in reference images while faithfully reconstructing the known regions.}
    \label{fig:dtu_qual}

\end{figure*}

\subsection{Experiment results}

\paragraph{Comparison with non-generative novel view synthesis models.} We compare our method with non-generative novel view synthesis models~\citep{charatan2024pixelsplat, chen2024mvsplat, ye2024no} on RealEstate10K~\citep{zhou2018stereo} using a challenging far-view setting that requires extensive inpainting of missing regions. We evaluate on three target views conditioned on two reference views, with target cameras positioned far from reference cameras to create large unknown areas. As shown in Table~\ref{table:nvs_quan}, our method outperforms state-of-the-art approaches even without camera pose access. Non-generative methods struggle in this extrapolative setting due to their inability to generate unseen regions, being limited to fusing existing input views. In contrast, our diffusion-based approach enables strong performance on both interpolation and extrapolation tasks. The qualitative results (Fig.~\ref{fig:qual_compare}) demonstrate semantically plausible inpainting and accurate geometry reconstruction, attributed to features that incorporate both geometric and semantic information.

\paragraph{Zero-shot evaluation.} 
We evaluate the generalization capability of our method using the DTU~\citep{jensen2014large} dataset, which was not seen during training. To comprehensively assess the generalization performance, we conduct evaluations under both near-view and far-view settings. For near-view, we follow the setting from MVSplat~\citep{chen2024mvsplat}, while the far-view setting is constructed by selecting the farthest view as the target. Table~\ref{table:dtu_zeroshot} shows that our method outperforms previous methods~\citep{charatan2024pixelsplat, chen2024mvsplat, ye2024no, zhang2025flare, jin2024lvsm} across both settings.
The qualitative results from Fig.~\ref{fig:dtu_qual} show that our method produces accurate geometry and semantically consistent inpainting, even in challenging target viewpoint of the out-of-domain data.

We also evaluate our approach using a single reference image against warping-and-inpainting approaches, LucidDreamer~\citep{chung2023luciddreamer}, GenWarp~\cite{seo2024genwarp}, and ViewCrafter~\cite{yu2024viewcrafter}. Evaluation is conducted on the DTU dataset~\citep{jensen2014large}. Table~\ref{table:dtu_zeroshot} demonstrates that our framework achieves superior performance in SSIM and LPIPS, maintaining competitive results in PSNR. 



\begin{figure*}
    \begin{center}
        \includegraphics[width=0.8\textwidth]{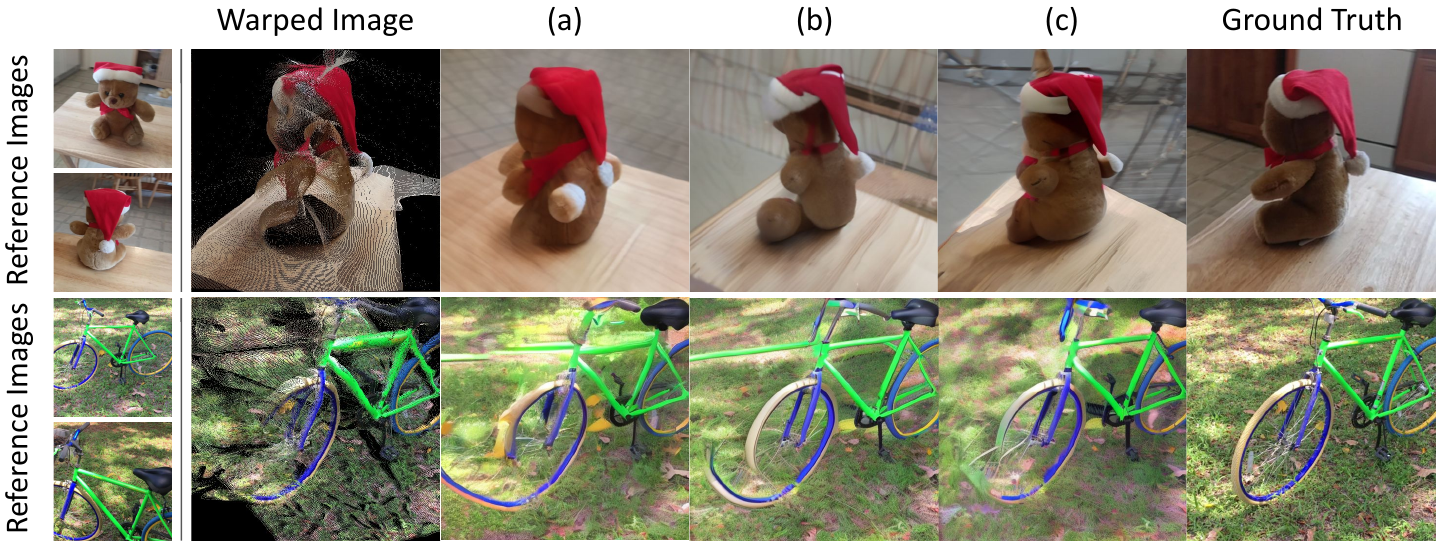}
    \end{center}
    \caption{
        \textbf{Qualitative results for ablation study.} \textbf{(Top):} Both (a) and (b) fail to reconstruct structurally consistent outputs, exhibiting misaligned body parts such as the arms, legs, and hat. In contrast, (c) effectively preserves both semantic consistency and structural integrity, producing coherent reconstructions aligned with the ground truth. \textbf{(Bottom):} Both (a) and (b) exhibit noticeable distortions in the wheel structure and fail to inpaint occluded background. Meanwhile, (c) achieves more accurate structural reconstruction and background inpainting, demonstrating superior semantic and geometric consistency.
    }
\label{fig:ablation_qual}
\vspace{-10pt}
\end{figure*}

\begin{table}[t!]
\begin{center}
    \centering
    \resizebox{1.0\linewidth}{!}{%
    \begin{tabular}{c|ccc}
        \toprule
         \textbf{Method} & PSNR$\uparrow$ & SSIM$\uparrow$ & LPIPS$\downarrow$  \\
        \midrule
         PixelSplat$^\dagger$~\citep{charatan2024pixelsplat}   
         & 14.01 & 0.582 & 0.384   \\
         MVSplat$^\dagger$~\citep{chen2024mvsplat}  
         & 12.13 & 0.534 & 0.380 \\
         NopoSplat~\citep{ye2024no}  
         & 14.36 & 0.538 & 0.389  \\
        \midrule
         \textbf{ReNoV w/ VGGT (Ours)} & \textbf{17.49} & \textbf{0.598} & \textbf{0.247}\\
        \bottomrule
    \end{tabular}}
\caption{\textbf{In-domain evaluation for a far-view setting}. We provide a quantitative analysis against prior feedforward methods using the in-domain Realestate10k~\citep{zhou2018stereo} dataset. $^\dagger$ denotes methods that require camera poses of the reference images.} 
\vspace{-10pt}
\label{table:nvs_quan} 

\end{center}
\end{table}


\subsection{Ablation} 
We explore how semantic and geometric conditioning features affect the performance of novel view synthesis. Specifically, we evaluate three configurations: (a) Baseline, utilizing semantic information from reference views via aggregated attention only; (b) Baseline with explicit geometric conditioning using predicted pointmaps; and (c) our final model conditioned on implicit semantic and geometric information by VGGT features. Quantitatively, Table~\ref{tab:ablation_quan} shows that explicit geometry conditioning through pointmaps in (b) improves overall performance compared to the baseline. Furthermore, conditioning VGGT features in (c) results in significant performance gains, highlighting the effectiveness of implicit geometric and semantic conditioning for extrapolative synthesis.

\begin{table}[t!]
\begin{center}
    \resizebox{0.5\textwidth}{!}{%
    \begin{tabular}{l|ccc}
    \toprule
          Components & PSNR$\uparrow$ & SSIM$\uparrow$ & LPIPS$\downarrow$ \\
          \midrule
          (a) Warped image only & 16.55 & 0.559 & 0.260 \\
          (b) Warped image + Pointmap condition & 16.93 & 0.594 & \textbf{0.243} \\
          (c) Pointmap condition + VGGT Feature & \textbf{17.50} & \textbf{0.598} & 0.247 \\ 
    \bottomrule    \end{tabular}}
    \caption{\textbf{Quantitative results for ablation study at Realestate10k dataset.} Evaluation results shows that leveraging the pointmaps and VGGT features enhances novel view synthesis performance.}
\label{tab:ablation_quan}
\vspace{-10pt}
\end{center}
\end{table}

In the qualitative evaluation (Fig.~\ref{fig:ablation_qual}), the baseline model (a) exhibits clear limitations in synthesizing structurally coherent novel views, resulting in perceptually distorted shapes and inconsistent reconstructions. Although explicit pointmap conditioning in (b) reduces geometric distortions, it still suffer from inaccurate inpainting due to insufficient semantic guidance. In contrast, our final configuration (c) utilizes VGGT features, which implicitly encode both semantic and geometric correspondences. This integrated conditioning allows the model to learn semantically consistent inpainting in challenging occluded regions, as well as structurally aligned reconstruction.

\begin{table}[t!]
\begin{center}
    \centering
    \resizebox{\linewidth}{!}{%
    \begin{tabular}{c|ccc}
    \toprule
          Removal percentage & PSNR$\uparrow$ & SSIM$\uparrow$ & LPIPS$\downarrow$ \\
          \midrule
          No removal  (Original setting) & 12.04 & 0.509 & 0.371 \\
          30\% removal & 11.89 & 0.507 & 0.366 \\
          50\% removal & 11.92 & 0.507 & 0.367 \\
    \bottomrule    
    \end{tabular}}
    \caption{\textbf{Robustness to degraded point clouds on DTU dataset.} Our method's performance remains stable even with geometric points used for novel view warping randomly masked.}
\label{tab:point_removal}
\end{center}
\vspace{-10pt}
\end{table}

\subsection{Robustness to Degraded Geometry}
\label{sec:robustness}

To evaluate our model's robustness to imperfect geometric inputs, we conducted ablation experiments by randomly subsampling point clouds, removing 30--50\% of points before warping-and-inpainting. We evaluate on the DTU dataset using the extreme (extrapolative) view setting, where point cloud projection is most susceptible to errors.

As shown in Table~\ref{tab:point_removal}, our model maintains stable performance despite significantly degraded inputs (PSNR drops only 0.12 with 50\% removal). This demonstrates that our diffusion framework effectively compensates for incomplete geometric information through learned generative priors and robust denoising capabilities, handling noisy or incomplete warped features inherently through its training process.



%% file: Writing/7_conclusion.tex
\section{Conclusion}

We introduce a diffusion‐based novel‐view synthesis framework that leverages VGGT’s multi‐view geometry features to unify precise reconstruction and semantically coherent inpainting. By reformulating synthesis as a warping‐and-inpainting task and injecting VGGT features into a conditioned diffusion U-Net, our method achieves state-of-the-art fidelity on both visible and occluded regions, outperforming existing diffusion-based approaches across standard benchmarks. These results underscore the value of rich geometric priors in guiding generative models, and open avenues for future extensions toward dynamic scenes and real-time applications.

%% file: appendix.tex







\appendix
\clearpage
\newpage
\section*{\Large Appendix}

In Sec.~\ref{supp:details}, we provide additional implementation details for our proposed method. In Sec.~\ref{supp:analy}, we present the results of additional analysis experiments to validate our approach. In Sec.~\ref{supp:exps}, we provide additional comparison to other baselines as well as additional qualitative ablation results and analysis.



\section{Additional details}
\label{supp:details}
\subsection{Training details}
In our training procedure, we initialize the image denoising U-Net from the Stable Diffusion 2.1 model and fine-tune it on a combination of large-scale datasets including RealEstate10K~\cite{zhou2018stereo}, Co3D~\cite{reizenstein21co3d}, and MVImgNet~\cite{yu2023mvimgnet}. The reference networks, which are architecturally identical to the image denoising U-Net (albeit without timestep embeddings), share the same initial weights and are trained solely to extract high-level semantic features from the input images. Ground-truth geometry is generated using an off-the-shelf geometry predictor, and only pointmaps from selected reference views are used during training for warping and proximity-based mesh conditioning. This strategy ensures that our model learns to synthesize both image and geometric representations in a mutually reinforcing manner. All models are trained with a batch size of 6 using two NVIDIA RTX A6000 GPUs (48GB) for a total of 60k training iterations.

To further stabilize training, we perform cross-modal attention instillation in a one-on-one fashion before combining the networks for joint training. This separate instillation phase allows the image and geometry branches to initially learn robust representations independently. Later, during simultaneous training, the geometry networks benefit from the deterministic cues provided by the image denoising network, which significantly improves consistency in geometry prediction. Our training schedule includes careful hyperparameter tuning, data augmentation, and regularization to mitigate overfitting while ensuring that the network generalizes well to unseen viewpoints.

\subsection{Evaluation details regarding test-time optimization} 
\label{supp:eval_detail}
NoPoSplat~\cite{ye2024no} and LVSM~\cite{jin2024lvsm} perform test-time optimization (TTO) of the target camera pose during evaluation. Specifically, these methods iteratively optimize the target camera extrinsic parameters by minimizing the reconstruction error between the rendered novel view and the ground-truth target image, as described in their respective papers. Notably, the test-time optimization directly minimizes the mean squared error (MSE) loss used for PSNR computation and relies on access to the ground-truth target image. 

This raises several concerns: (1) performance becomes highly sensitive to optimization hyperparameters (particularly iteration steps), which are often not explicitly specified, compromising reproducibility; (2) since the optimization objective directly aligns with the evaluation metric, the process may prioritize metric maximization over geometric accuracy—our analysis reveals cases where camera alignment with ground truth is sacrificed to minimize optimization loss; and (3) inference time increases substantially (e.g., from 3.1 to 14.24 seconds when extending from 100 to 800+ optimization steps). Therefore, for a fair comparison, we report metrics without test-time optimization in our main table (Tab.\ref{table:dtu_zeroshot}), as these more directly reflect learned model capabilities with deterministic and reproducible results. For completeness, we additionally report the performance of test-time optimization under the near-view setup in Tabs.\ref{tab:noposplat_tto} and \ref{tab:flare_tto}.

\begin{table}[t]
  \centering
  \small
  \caption{Effect of test-time camera pose optimization (TTO) in NoPoSplat on the DTU dataset.}
  \label{tab:noposplat_tto}
  \setlength{\tabcolsep}{6pt}
  \resizebox{\linewidth}{!}{%
    \begin{tabular}{l|ccc}
      \toprule
      Method & PSNR($\uparrow$) & SSIM($\uparrow$) & LPIPS($\downarrow$) \\
      \midrule
      NoPoSplat (Feed-forward output) 
      & 11.44 & 0.357 & 0.576 \\
      NoPoSplat + 100 optim.\ steps
      & 14.05 & 0.414 & 0.503 \\
      NoPoSplat + 200 optim.\ steps 
      & 15.89 & 0.478 & 0.419 \\
      NoPoSplat + 400 optim.\ steps 
      & 17.15 & 0.553 & 0.363 \\
      NoPoSplat + 800 optim.\ steps 
      & {17.22} & 0.558 & 0.356 \\
      \bottomrule
    \end{tabular}%
  }
\end{table}

\begin{table}[t]
  \centering
  \small
  \caption{Effect of test-time optimization (TTO) in FLARE on the DTU dataset.}
  \label{tab:flare_tto}
  \setlength{\tabcolsep}{8pt}
  \resizebox{\linewidth}{!}{%
    \begin{tabular}{l|ccc}
      \toprule
      Method & PSNR($\uparrow$) & SSIM($\uparrow$) & LPIPS($\downarrow$) \\
      \midrule
      FLARE (Feed-forward output) 
      & 13.25 & 0.381 & 0.502 \\
      FLARE + 100 optim.\ steps 
      & {17.35} & {0.588} & {0.298} \\
      \bottomrule
    \end{tabular}%
  }
\end{table}

\section{Additional Analysis}
\label{supp:analy}

\begin{figure*}[ht!]
    \centering
    \includegraphics[height=0.94\textheight]{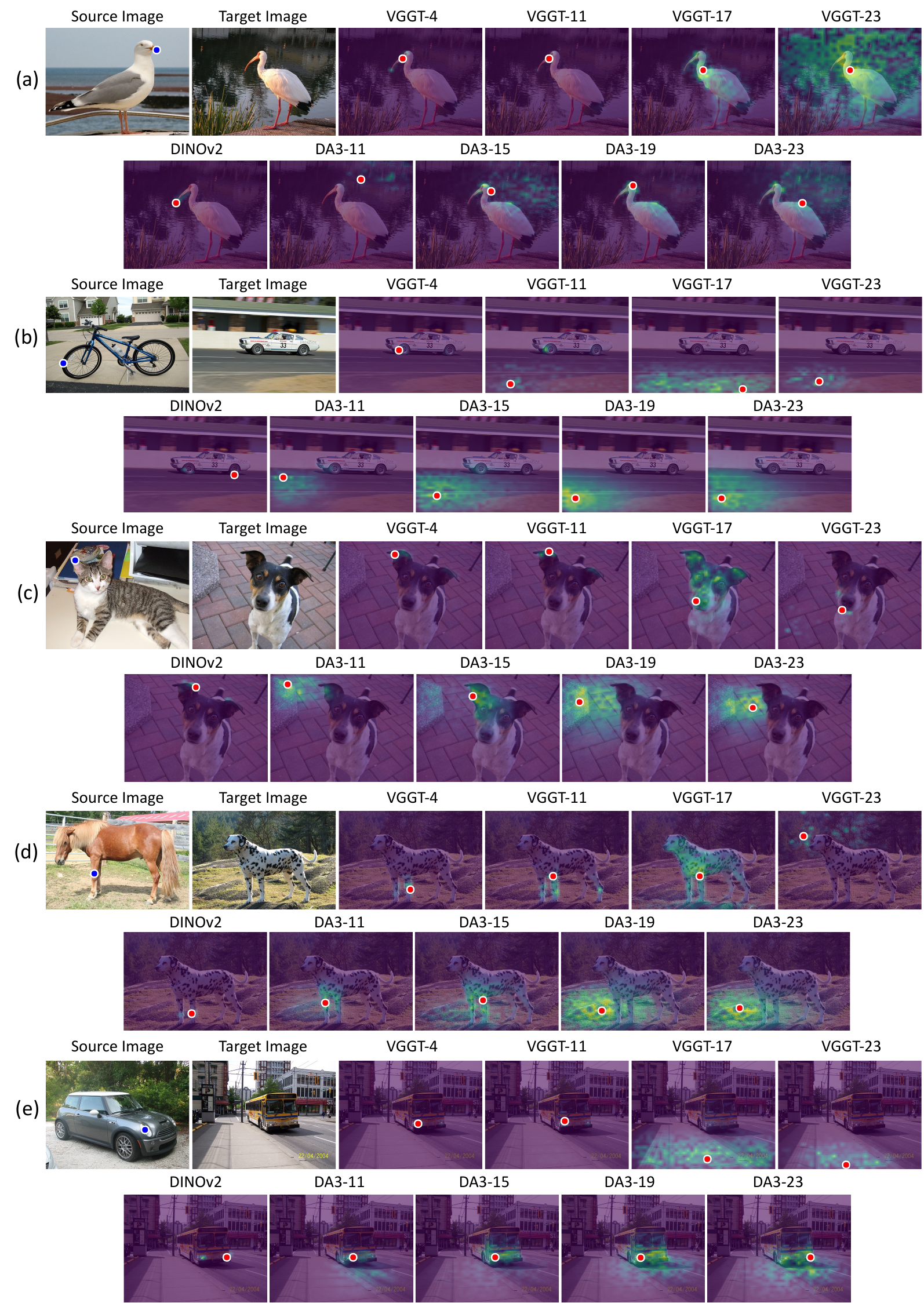}
    \caption{\textbf{Visualization of feature similarity map.} The top-leftmost figure shows the source image with a query point (blue dot), followed by the target image. Cosine similarity is computed between the query and all target patch features to assess semantic encoding. Early VGGT layers (4\textsuperscript{th}, 11\textsuperscript{th}) retain strong semantic signals, effectively highlighting fine-grained regions (e.g., beak, wheel, ear, etc.). DINOv2 captures rich semantics but with less precise localization. DA3 fails to capture meaningful semantic cues.}
    \label{fig:sem_corr_qual}
\end{figure*}

\subsection{Semantic Correspondence}




Additional qualitative results are presented in Figure~\ref{fig:sem_corr_qual}, further illustrating that the early layers of VGGT~\cite{wang2025vggt} encode rich semantic information, which gradually diminishes in deeper layers. These early layers also exhibit an ability to capture geometrically consistent semantics. For instance, in Figure~\ref{fig:sem_corr_qual}(e), when the query point is placed on the right headlight of a car, VGGT accurately identifies the corresponding right headlight in the target image. In contrast, DINOv2~\cite{oquab2023dinov2} matches the left light, ignoring spatial alignment, while DA3~\cite{lin2025depth} produces sparse and imprecise correspondences, often highlighting regions that are weakly related to the underlying semantics. Similar patterns appear throughout Figure~\ref{fig:sem_corr_qual}, where early VGGT layers demonstrate direction-aware and spatially accurate semantic matching, often outperforming DINOv2 in both precision and structure-awareness.

\begin{figure*}[ht!]
    \centering
    \includegraphics[width=0.73\linewidth]{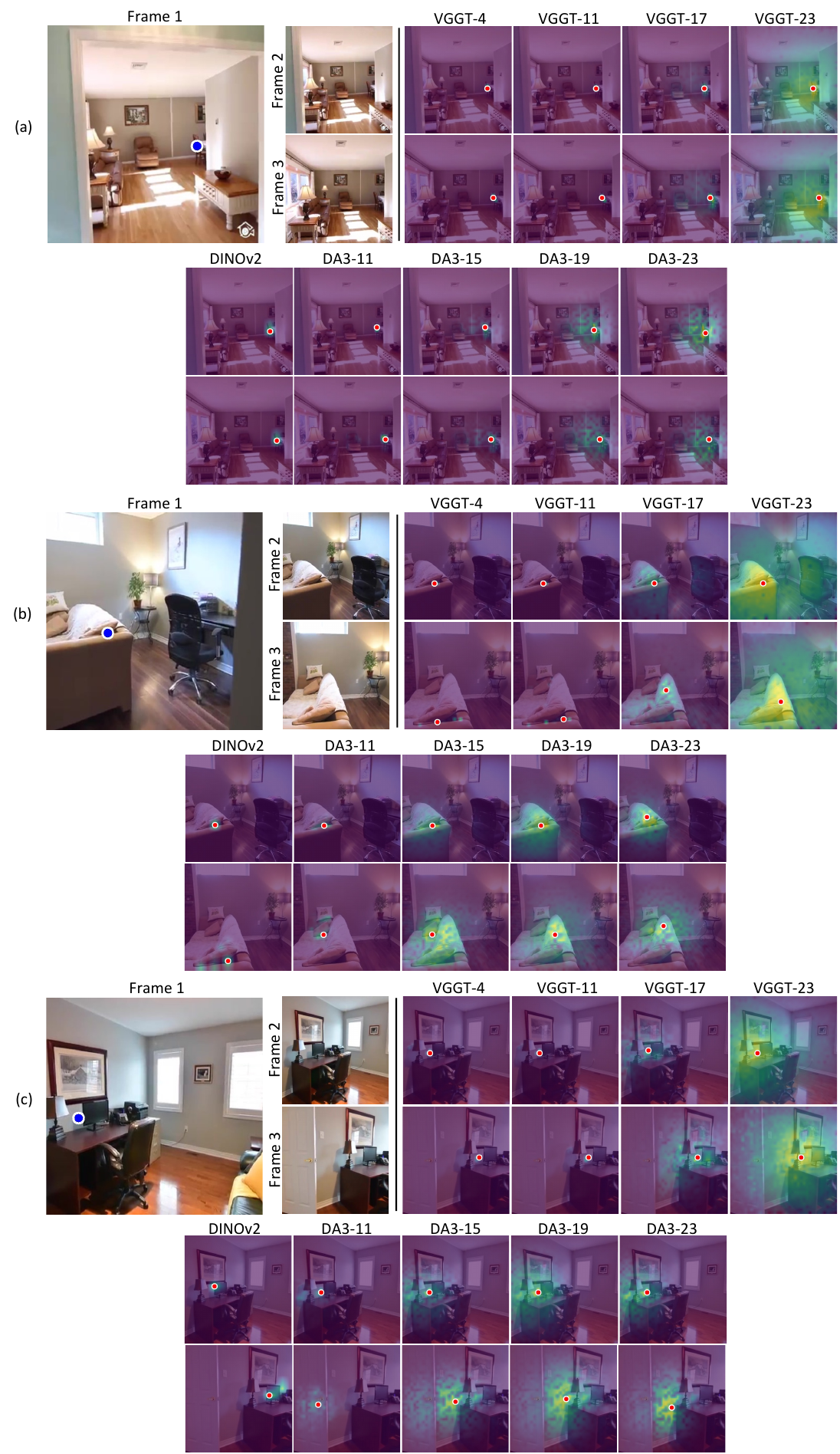}
     \caption{\textbf{Geometric correspondence evaluation.} A query point (blue dot) is selected in Frame 1, and cosine similarity maps are computed in Frame 2 and Frame 3. All layers of VGGT accurately identify the correct object aligned with the query point. In contrast, DA3 successfully localizes the object in the nearby view (Frame 2) but fails in the distant view (Frame 3). This illustrates that deeper layers of VGGT capture geometric structure more reliably than others.}
    \label{fig:geo_qual}
\end{figure*}

\subsection{Geometric Correspondence}
Additional qualitative results are presented in Figure~\ref{fig:geo_qual}, highlighting that the deeper layers of VGGT~\cite{wang2025vggt} better capture geometric structure. In contrast, DA3~\cite{lin2025depth} successfully localizes the object in the nearby view (Frame 2) but fails in the distant view (Frame 3). This observation is consistent with the quantitative results shown in Fig.~\ref{fig:main_analysis_a}, where VGGT achieves high scores across all layers, whereas DA3 exhibits low performance except at its peak layer.

\begin{figure*}[t]
    \centering
    \includegraphics[width=0.8\linewidth]{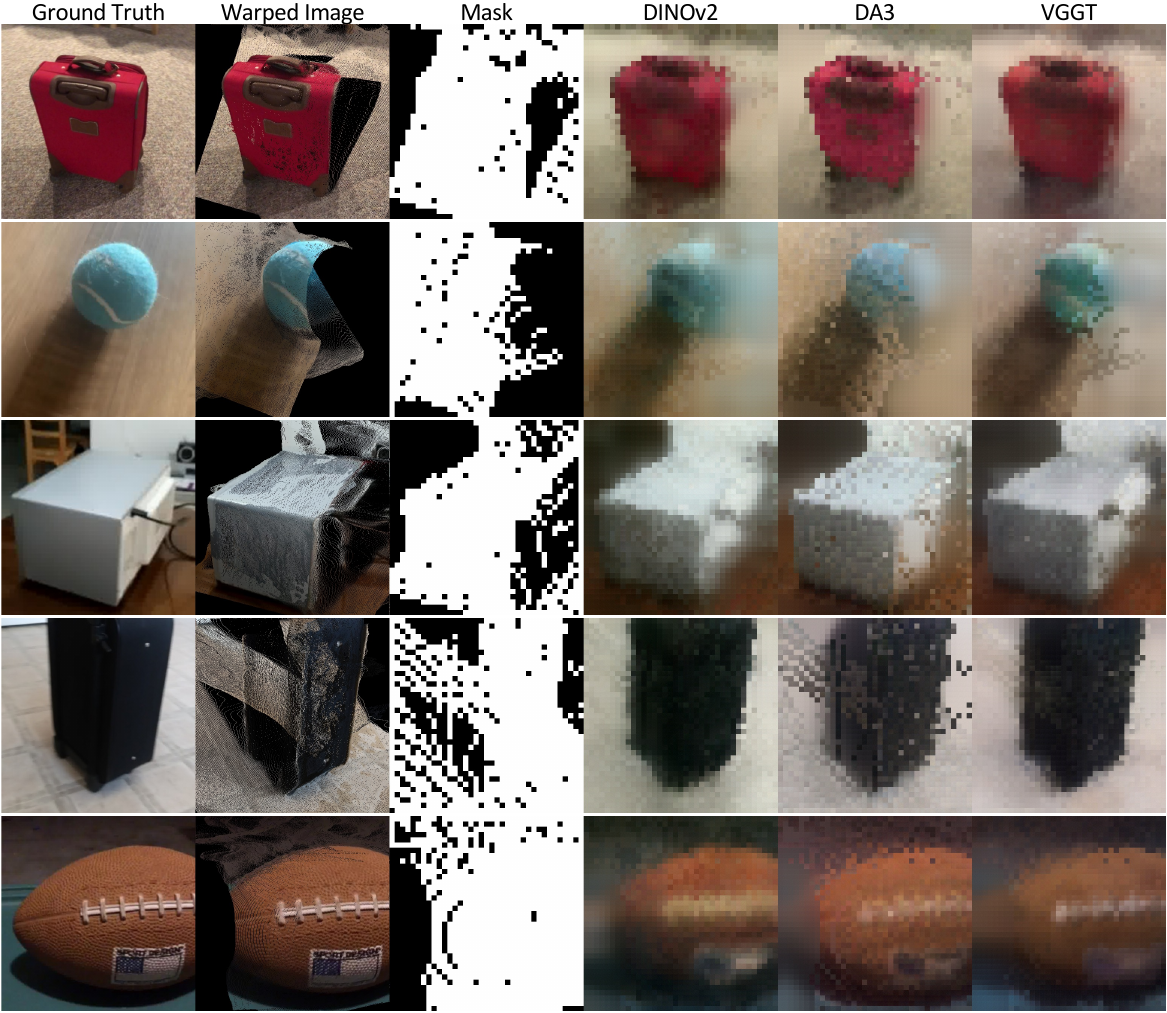}
    \caption{\textbf{Qualitative results of representation reconstruction.} The "Warped Image" column shows warped images, where features are warped using the same predicted pointmaps, resulting in corresponding feature-level holes.}
    \label{fig:appendix_recon_qual}
    \vspace{-10pt}
\end{figure*}

\subsection{Representation Reconstruction Probing}
To provide additional experimental validation of this hypothesis through systematic probing, we provide additional experimental results on probing, using a shallow MAE~\cite{he2022masked} decoder trained to predict target view images from warped reference view features, as given in the main paper. The additional experimental results from this probing analysis offer further empirical evidence supporting our feature representation choices and their effectiveness in the warping-and-inpainting framework.

\paragraph{Qualitative results.}
Fig.~\ref{fig:appendix_recon_qual} shows additional qualitative results for different features-DA3~\cite{lin2025depth}, DINOv2~\cite{oquab2023dinov2}, and VGGT~\cite{wang2025vggt}. Among them, images generated using VGGT features exhibit the highest geometric and semantic fidelity to the ground truth, highlighting VGGT's ability to effectively encode both multi-view geometric correspondences and rich semantic context.
\vspace{-10pt}

\begin{figure*}[t]
    \centering
    \includegraphics[width=1.0\linewidth]{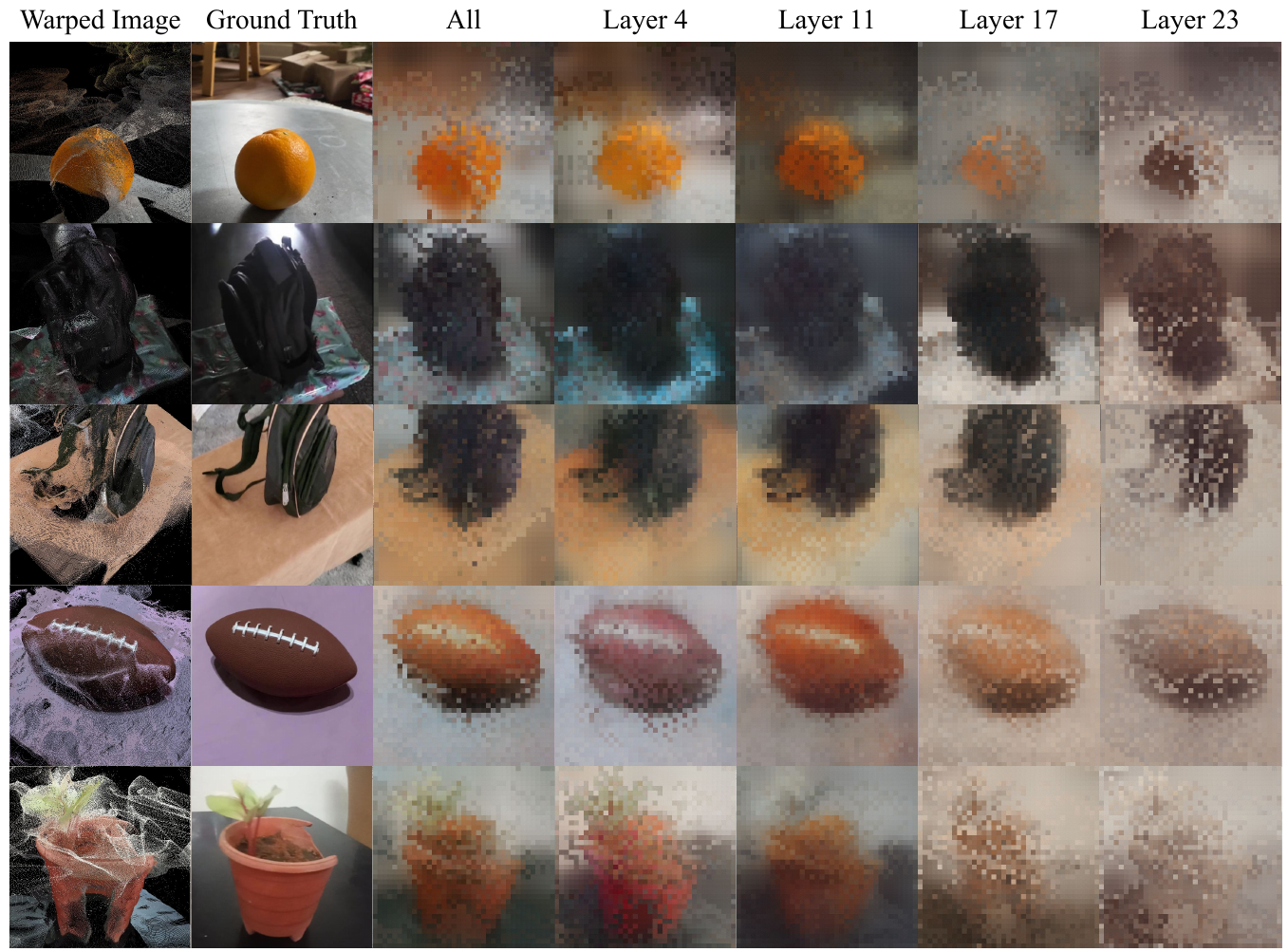}
    \caption{\textbf{Per-layer probing qualitative results.} We visualize generation results using VGGT features extracted from individual layers and their combination. Early layers (4\textsuperscript{th}, 11\textsuperscript{th}) retain rich semantic information, producing semantically coherent images with accurate color and texture. In contrast, deeper layers (17\textsuperscript{th}, 23\textsuperscript{rd}) emphasize geometric structure but lack semantic detail. Combining features across all layers yields the most faithful reconstructions, achieving both structurally accurate and semantically realistic outputs.
    }
    \label{fig:appendix_recon_abl_qual}
    \vspace{-10pt}
\end{figure*}

\paragraph{Ablation.}
We conduct an ablation study to investigate the representational capability of VGGT~\cite{wang2025vggt} features extracted from different layers. Specifically, we train a shallow MAE~\cite{he2022masked} decoder on features from the 4\textsuperscript{th}, 11\textsuperscript{th}, 17\textsuperscript{th}, and 23\textsuperscript{rd} layers, and evaluate their generation performance qualitatively. Fig.~\ref{fig:appendix_recon_abl_qual} demonstrates that deeper layers tend to capture more geometric structure but offer less semantic detail. In contrast, aggregating features across all layers results in the most visually plausible image, indicating effective reconstruction fidelity and more semantically coherent inpainting.




\section{Additional Results}
\label{supp:exps}

\subsection{Qualitative Results} 
\begin{figure*}[ht!]
    \centering
    \includegraphics[width=1.0\textwidth]{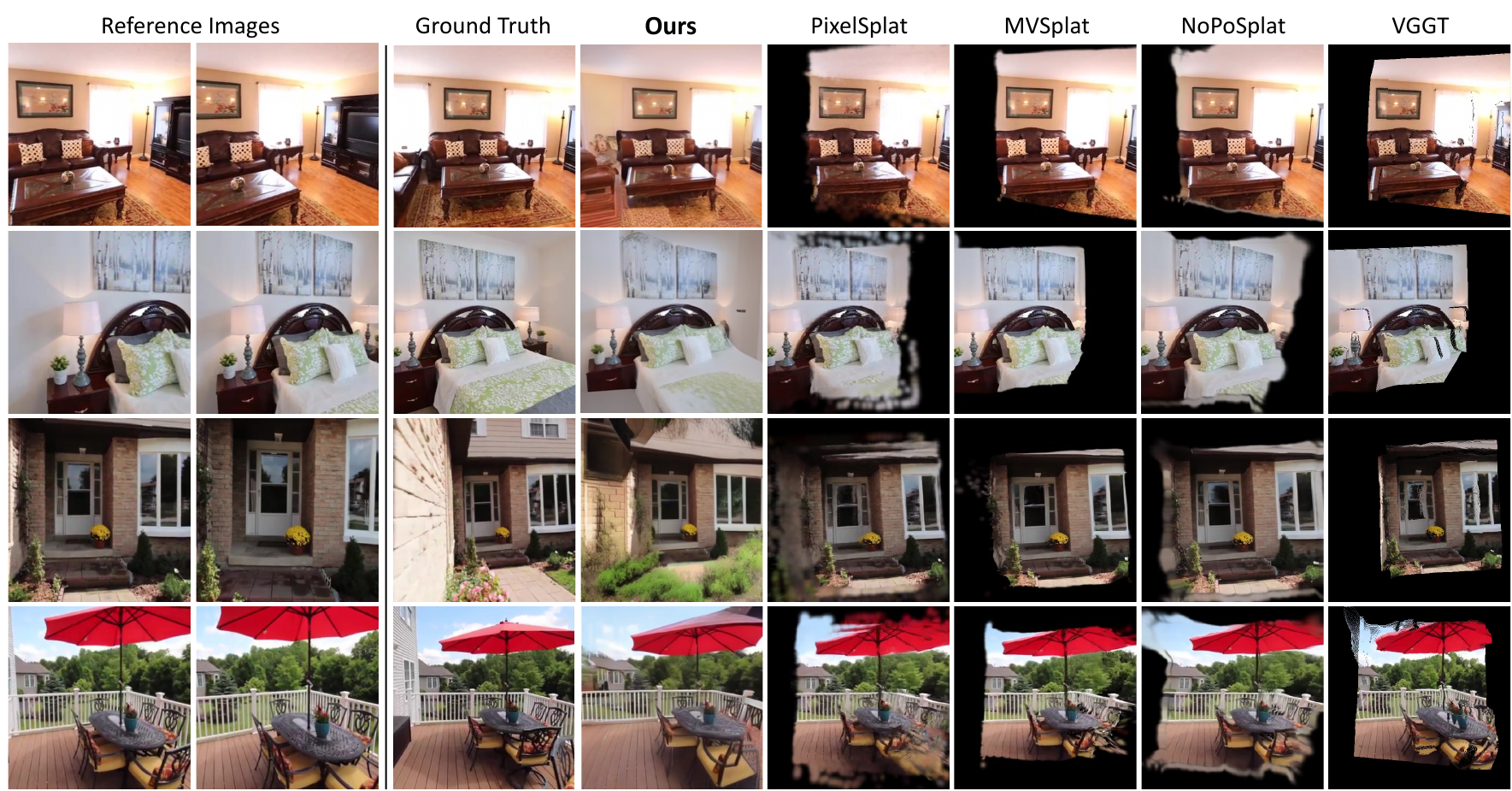}
    \caption{\textbf{Qualitative comparison on RealEstate10k.} 
    Qualitative results demonstrate the extrapolative capability of our model (ReNoV w/ VGGT) to plausibly generate locations not seen in the reference images, while faithfully reconstructing the known regions. }
    \label{fig:qual_compare}
    \vspace{-10pt}
\end{figure*}

Fig.~\ref{fig:qual_compare} presents qualitative comparisons on the RealEstate10k dataset. ReNoV w/ VGGT produces visually coherent reconstructions of the observed regions while plausibly extrapolating to unseen locations beyond the reference views. 
These results highlight the model’s ability to maintain spatial consistency and generate realistic scene content under view extrapolation.

\begin{figure*}[ht!]
    \centering
    \includegraphics[width=1.0\textwidth]{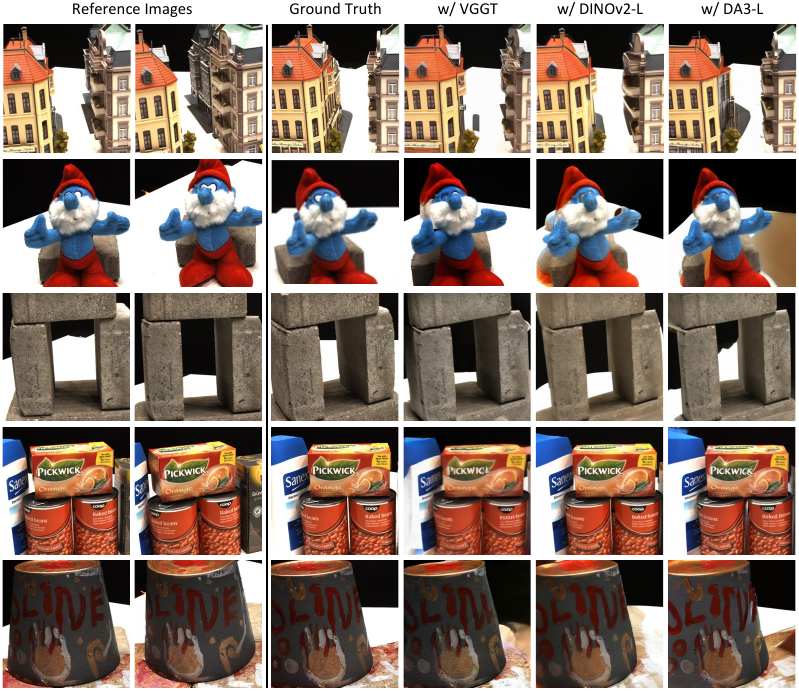}
    \caption{\textbf{Qualitative comparison of our framework across feature representations on DTU.} We visualize the robustness of our synthesis pipeline when utilizing different feature backbones, using the same scenes shown in Fig.~\ref{fig:dtu_qual}). Compared to state-of-the-art baselines, our method effectively mitigates structural artifacts and produces more coherent inpainting, regardless of the conditioning feature representation.
    }
    \label{fig:qual_dino_da3}
    \vspace{-10pt}
\end{figure*}

Fig. \ref{fig:qual_dino_da3} provides a qualitative results of our framework integrated with diverse feature representations, using the same scenes from the DTU dataset shown in Fig.~\ref{fig:dtu_qual} for a direct comparison. Our method consistently yields high-fidelity synthesis that exceeds the baselines~\cite{charatan2024pixelsplat, chen2024mvsplat, ye2024no, zhang2025flare, jin2024lvsm}. Notably, while baselines frequently suffer from geometric distortions or blurring in unobserved areas, our framework maintains strict multi-view consistency. Furthermore, it demonstrates a superior capacity for generative inpainting, producing perceptually plausible textures sacrificing structural integrity.

\subsection{Ablation Study}

\begin{figure*}[t]
    \centering
    \includegraphics[width=1.0\linewidth]{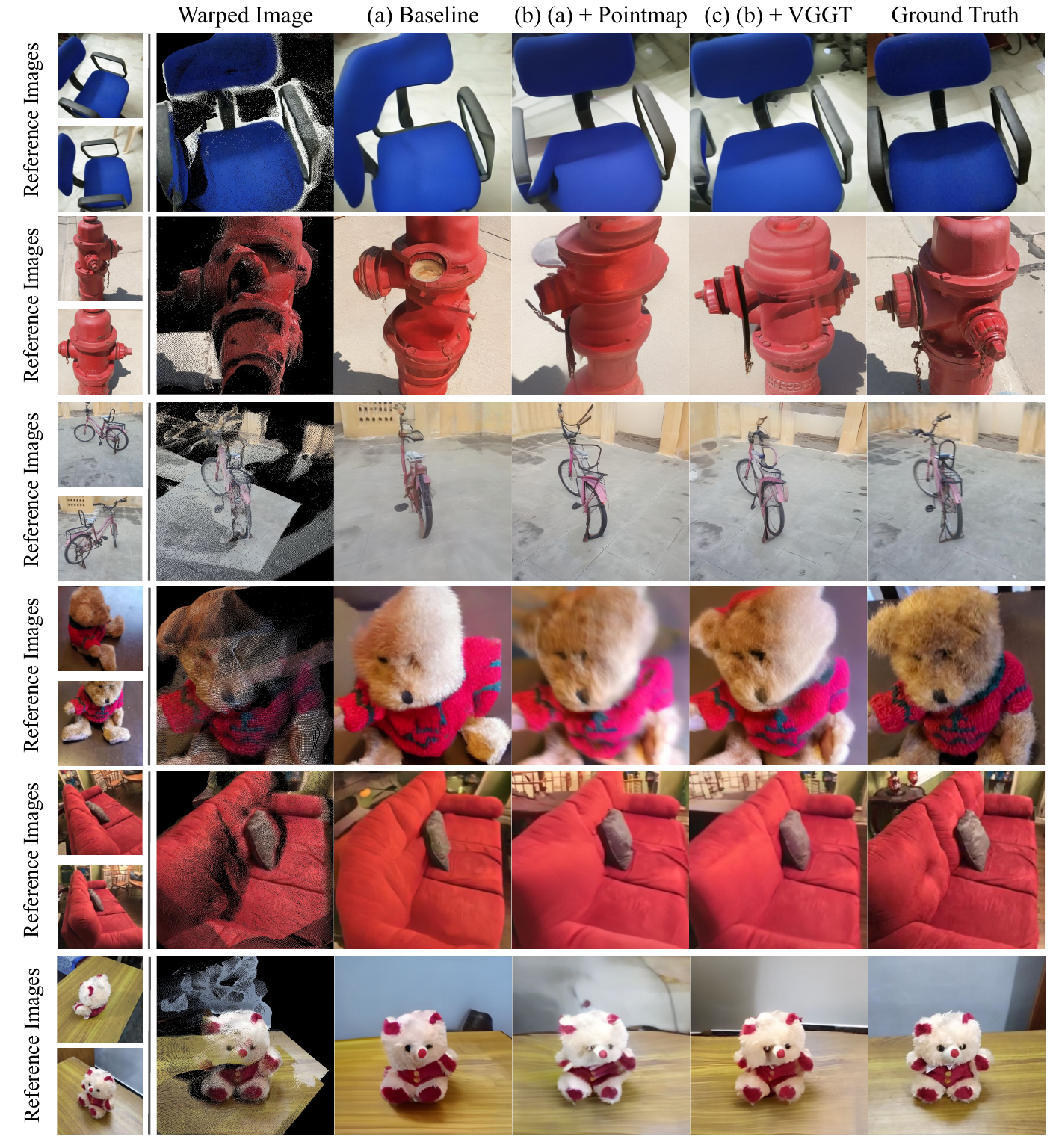}
    \caption{\textbf{Qualitative ablation results.} (a) \textit{Baseline}: Lacks geometric guidance, resulting in misaligned structures (e.g., distorted chair, missing bicycle wheel, and incomplete teddy bear). (b) \textit{Baseline + pointmap}: Improves geometric alignment but suffers from distortion due to noisy geometry and inaccurate inpainting (e.g., deformed chair seat). (c) \textit{Ours with VGGT features}: Implicit semantic and geometric conditioning enables accurate reconstruction of visible regions and plausible inpainting of occluded areas.}
    \label{fig:appendix_co3d_qual}
    \vspace{-10pt}
\end{figure*}

\paragraph{Qualitative results.}
Fig.~\ref{fig:appendix_co3d_qual} shows additional ablation results for three configurations: (a) baseline with semantic-only conditioning; (b) baseline + explicit geometric guidance via pointmaps; (c) ours with implicit semantic and geometric conditioning using VGGT~\cite{wang2025vggt} features. Conditioning on VGGT features enables the model to achieve more accurate reconstructions and plausible inpainting by leveraging rich implicit geometric and semantic information. In contrast, (a) and (b) exhibit noticeable geometric distortions and incomplete inpainting, highlighting the limitations of lacking or noisy geometric cues.

\begin{figure*}[t]
    \vspace{-20pt}
    \centering
    \includegraphics[width=0.8\linewidth]{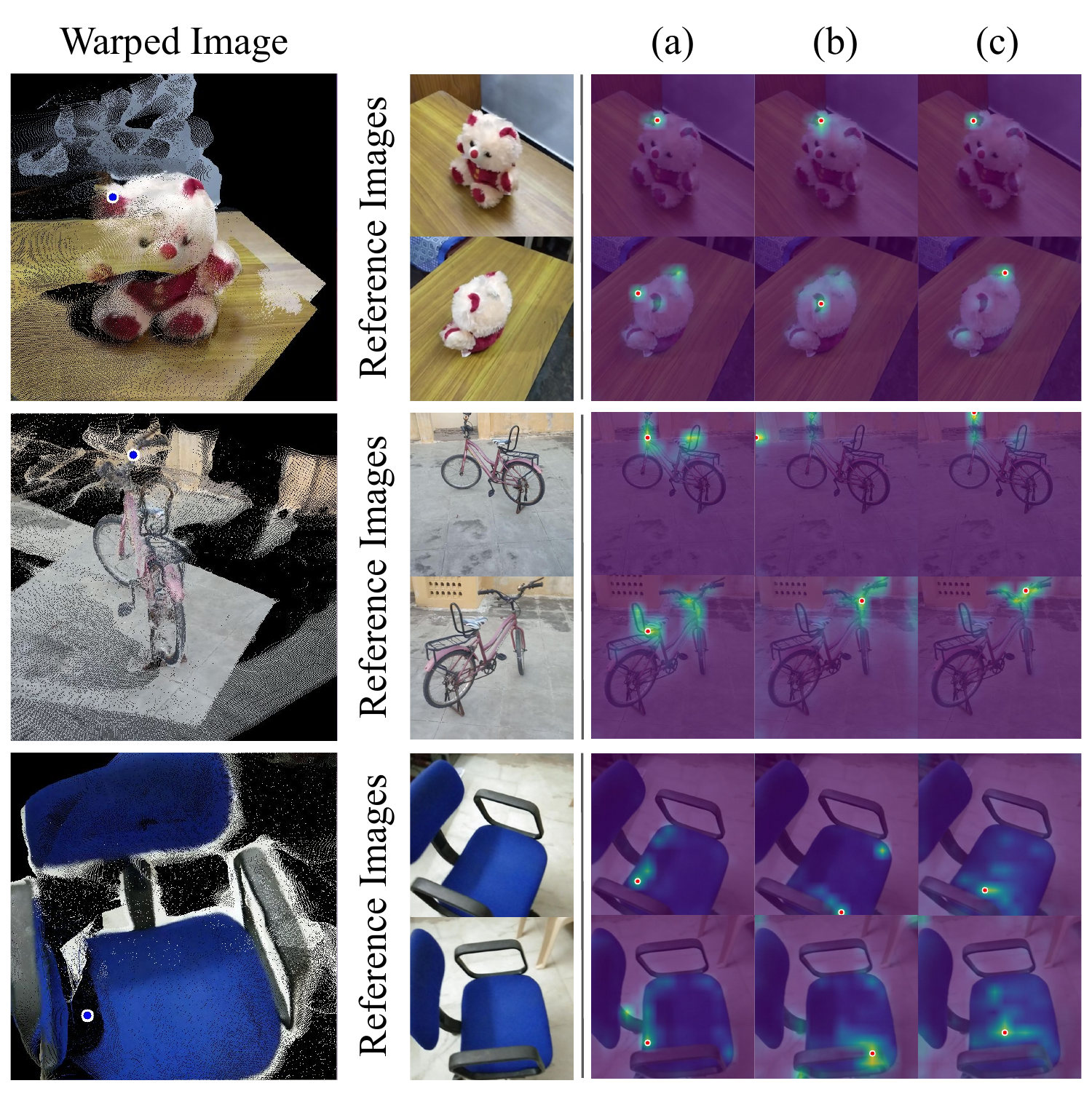}
    \caption{\textbf{Attention map visualization for ablation study.} The leftmost column shows a query point (blue dot) in the warped image, with corresponding cross-attention maps over reference images shown on the right. Configurations (a) and (b) attend to incorrect regions for both reconstruction (e.g., teddy bear’s ear, bicycle handle) and inpainting (e.g., chair seat), due to limited or noisy geometric guidance. In contrast, VGGT-based conditioning (c) guides attention to geometrically and semantically aligned regions, accurately distinguishing fine structures such as the correct ear of the teddy bear.}
    \label{fig:appendix_attn_vis}
    \vspace{-10pt}
\end{figure*}

\paragraph{Attention map visualization.}
We further analyze the cross-view attention maps of the denoising U-Net trained under configurations (a), (b), and (c). As shown in Fig.~\ref{fig:appendix_attn_vis}, the baseline model (a) attends to geometrically and semantically misaligned regions in the reference images, leading to inaccurate reconstruction and inpainting. Explicit geometric guidance via pointmaps (b) partially reduces this misalignment but remains insufficient due to noisy and incomplete geometric correspondences. In contrast, our final model conditioned on VGGT~\cite{wang2025vggt} features (c) accurately attends to geometrically and semantically consistent regions in the reference views, significantly enhancing the quality of synthesized images. This confirms that VGGT features effectively guide cross-view attention toward optimal reference positions by implicitly encoding comprehensive geometric and semantic correspondences.

\clearpage
\newpage

